\documentclass{article}

\PassOptionsToPackage{numbers,sort}{natbib}

\usepackage[preprint]{neurips_2022}

\usepackage[utf8]{inputenc} 
\usepackage[T1]{fontenc}    
\usepackage[hidelinks]{hyperref}       
\usepackage{url}            
\usepackage{booktabs}       
\usepackage{amsfonts}       
\usepackage{nicefrac}       
\usepackage{microtype}      
\usepackage{xcolor}         
\usepackage{lipsum}

\usepackage{natbib}
\usepackage{amsmath}
\usepackage{amssymb}
\usepackage{mathtools}
\usepackage{amsthm}
\usepackage{mathrsfs}
\usepackage{thmtools}
\usepackage{xspace}
\usepackage{dsfont}
\usepackage{wrapfig}
\usepackage{multirow}
\usepackage{bibunits}

\defaultbibliography{neurips_2022}
\defaultbibliographystyle{plainnat}

\theoremstyle{plain}
\newtheorem{theorem}{Theorem}[section]

\theoremstyle{definition}
\newtheorem{definition}[theorem]{Definition}

\theoremstyle{remark}

\definecolor{kqcolor}{RGB}{219,90,107}

\newcommand{\laprep}{{\fontfamily{cmss}\selectfont{LapRep}}\xspace}
\newcommand{\ralaprep}{{\fontfamily{cmss}\selectfont{RA-LapRep}}\xspace}

\makeatletter
\DeclareRobustCommand\onedot{\futurelet\@let@token\@onedot}
\def\@onedot{\ifx\@let@token.\else.\null\fi\xspace}

\def\eg{\emph{e.g}\onedot} 
\def\ie{\emph{i.e}\onedot}

\def\wrt{w.r.t\onedot} 

\makeatother

\DeclareMathOperator*{\argmax}{arg\,max}

\title{Reachability-Aware Laplacian Representation\\ in Reinforcement Learning}

\author{%
  Kaixin Wang\textsuperscript{1}
  \And
  Kuangqi Zhou\textsuperscript{2}
  \And
  Jiashi Feng\textsuperscript{3}
  \And
  Bryan Hooi\textsuperscript{1,4}
  \And
  Xinchao Wang\textsuperscript{1,2}
  \AND
  \texttt{\{kaixin.wang, kzhou\}@u.nus.edu} \\
  \texttt{jshfeng@gmail.com} \\
  \texttt{\{dcsbhk, xinchao\}@nus.edu.sg}
  \AND
  \textnormal{\textsuperscript{1}Institute of Data Science, National University of Singapore} \\
  \textsuperscript{2}Electrical and Computer Engineering,
  National University of Singapore\\
  \textsuperscript{3}ByteDance\\
  \textsuperscript{4}School of Computing,
  National University of Singapore
}

\begin{document}
\begin{bibunit}

\maketitle

\begin{abstract}
In Reinforcement Learning (RL), Laplacian Representation (\laprep) is a task-agnostic state representation that encodes the geometry of the environment.
A desirable property of \laprep stated in prior works is that the Euclidean distance in the \laprep space roughly reflects the reachability between states, which motivates the usage of this distance for reward shaping.
However, we find that \laprep does not necessarily have this property in general: two states having small distance under \laprep can actually be far away in the environment.
Such mismatch would impede the learning process in reward shaping.
To fix this issue, we introduce a Reachability-Aware Laplacian Representation (\ralaprep), by properly scaling each dimension of \laprep.
Despite the simplicity, we demonstrate that \ralaprep can better capture the inter-state reachability as compared to \laprep, through both theoretical explanations and experimental results.
Additionally, we show that this improvement yields a significant boost in reward shaping performance and also benefits bottleneck state discovery.
\end{abstract}

\section{Introduction}
Reinforcement Learning (RL) seeks to learn a decision strategy that advises the agent on how to take actions according to the perceived states~\cite{sutton2018reinforcement}.
The state representation plays an important role in the agent's learning process \textemdash~ a proper choice of the state representation can help improve generalization~\cite{zhang2018decoupling, stooke2020decoupling, agarwal2021contrastive}, encourage exploration~\cite{pathak2017curiosity, machado2017laplacian, machado2020count} and enhance learning efficiency~\cite{dubey2018investigating, wu2018laplacian, wang2021towards}. In studying the state representation, one direction of particular interest is to learn task-agnostic representation that encodes transition dynamics of the environment~\cite{mahadevan2007proto, machado2021temporal}.

Along this line, the Laplacian Representation (\laprep) has received increasing attention~\cite{mahadevan2005proto,machado2017laplacian, wu2018laplacian,wang2021towards,erraqabi2022temporal}.
Specifically, the \laprep is formed by the $d$ smallest eigenvectors of the Laplacian matrix of the graph induced from the transition dynamic (see Section~\ref{sec:bg-laprep} for definition).
It is assumed in prior works~\cite{wu2018laplacian, wang2021towards} that \laprep has a desirable property: the Euclidean distance in the \laprep space roughly reflects the \emph{reachability} among states~\cite{wu2018laplacian, wang2021towards}, \ie, smaller distance implies that it is easier to reach one state from another.
This motivates the usage of the Euclidean distance under \laprep for reward shaping~\cite{wu2018laplacian, wang2021towards}.

\begin{wrapfigure}[15]{hr}{0.45\textwidth}
    \centering
    \vspace{-10pt}
    \includegraphics[width=0.45\textwidth]{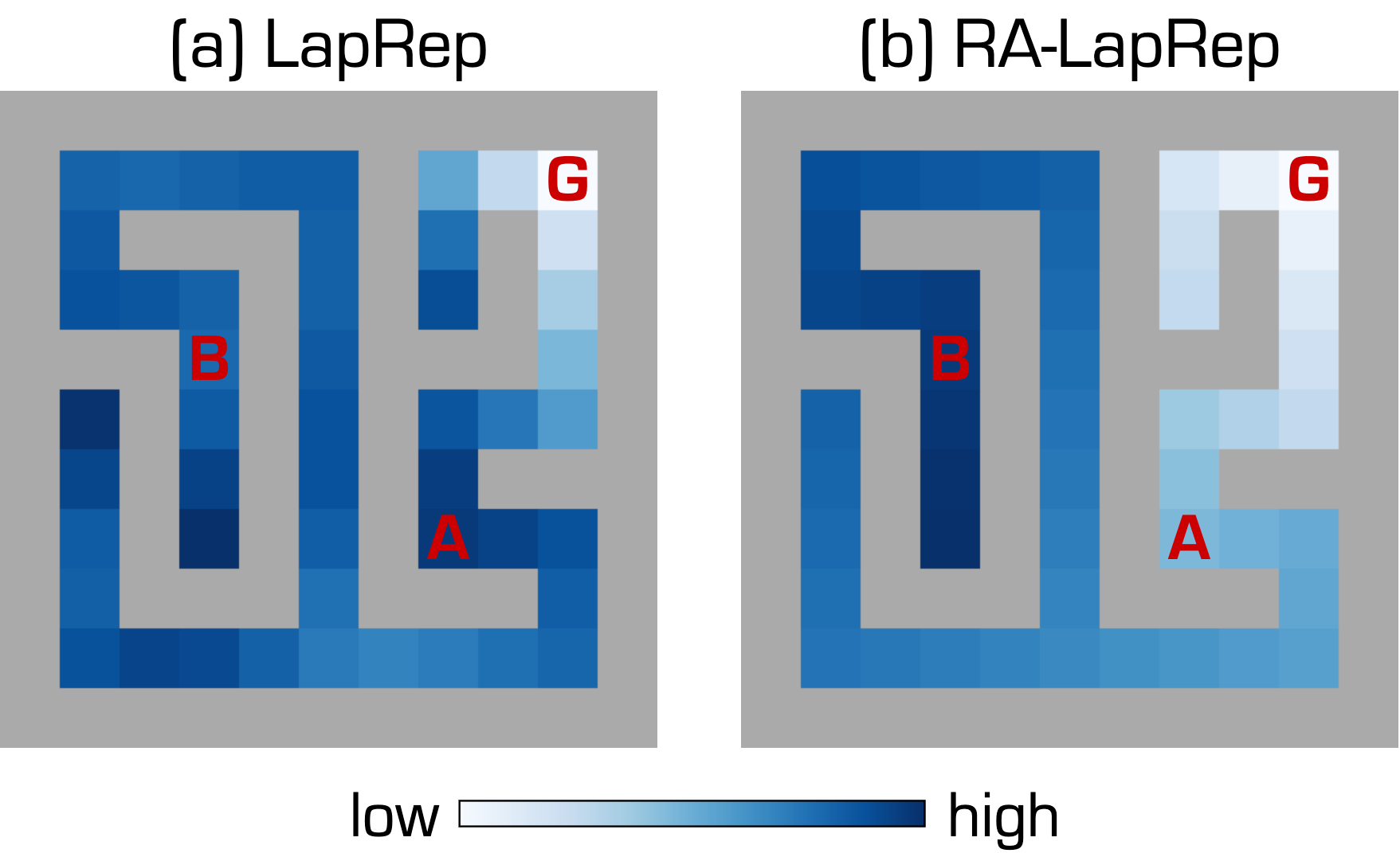}
    \caption{Euclidean distances between each state and the goal state \texttt{G}, under \laprep \textbf{(a)} and \ralaprep \textbf{(b)}.}
    \label{fig:intro}
\end{wrapfigure}

However, there lacks formal justification in previous works~\cite{wu2018laplacian, wang2021towards} for this property.
In fact, it turns out that the Euclidean distance under \laprep does \emph{not} correctly capture the inter-state reachability in general.
Figure~\ref{fig:intro}~(a) shows an example.
Under \laprep, a state that has larger distance (\eg, \texttt{A}) might actually be closer to the goal than another state (\eg, \texttt{B}).
Consequently, when the agent moves towards the goal, the pseudo-reward provided by \laprep would give a wrong learning signal.
Such mismatch would hinder the learning process with reward shaping and result in inferior performance.

In this work, we introduce a Reachability-Aware Laplacian Representation (\ralaprep) that reliably captures the inter-state distances in the environment geometry (see Figure~\ref{fig:intro}~(b)).
Specifically, \ralaprep is obtained by scaling each dimension of the \laprep by the inverse square root of the corresponding eigenvalue.
Despite its simplicity, \ralaprep has theoretically justified  advantages over \laprep in the following two aspects.
First, the Euclidean distance under \ralaprep can be interpreted as the \textit{average commute time}, which measures the expected number of steps required in a random walk to navigate between two states.
Thus, such distance provides a good measure of the reachability.
In contrast, to our best knowledge, there lacks a connection between the Euclidean distance under \laprep and the reachability.
Second, \ralaprep is equivalent to the embedding computed by the classic multidimensional scaling (MDS)~\cite{BorgGroenen2005}, which preserves pairwise distances globally~\cite{tenenbaum2000global}.
\laprep, on the other hand, preserves only local information (\ie, mapping neighboring states close), since it is essentially Laplacian eigenmap~\cite{belkin2003laplacian}.
Thus, \laprep is inherently incompetent for measuring the inter-state reachability.

To further validate the advantages of \ralaprep over \laprep, we conduct experiments to compare them on two discrete gridworld and two continuous control environments.
The results show that \ralaprep indeed performs much better in capturing the inter-state reachability as compared to \laprep.
Furthermore, when used for reward shaping in the goal-reaching tasks, \ralaprep significantly outperforms \laprep.
In addition, we show that \ralaprep can be used to discover the bottleneck states based on graph centrality measure, and more accurately find the key states than \laprep.

The rest of the paper is organized as follows. In Section~\ref{sec:background}, we give some background about RL and \laprep in RL. In Section~\ref{sec:method}, we introduce the new \ralaprep, explain why it is more desirable than \laprep, and discuss how to approximate it with neural networks in environments with a large or continuous state space. Then, we conduct
experiments to demonstrate the advantages of \laprep over \ralaprep in Section~\ref{sec:exp}. In Section~\ref{sec:related},
we review related works and Section~\ref{sec:conclusion} concludes the paper.



\section{Background}
\label{sec:background}

\emph{Notations}. We use boldface letters (\eg, $\mathbf{u}$) for vectors, and calligraphic letters (\eg, $\mathcal{U}$) for sets. 
For a vector $\mathbf{u}$, $\lVert\mathbf{u}\rVert$ denotes its $L_2$ norm, $\mathrm{diag}(\mathbf{u})$ denotes a diagonal matrix whose main diagonal is $\mathbf{u}$.
We use $\mathbf{1}$ to denote an all-ones vector, whose dimension can be inferred from the context.

\subsection{Reinforcement Learning}
In the Reinforcement Learning (RL) framework~\cite{sutton2018reinforcement}, an agent aims to learn a strategy that advise how to take actions in each state, with the goal of maximizing the expected cumulative reward. We consider the standard Markov Decision Process (MDP) setting~\cite{puterman1990markov}, and describe an MDP with a quintuple $(\mathcal{S},\mathcal{A}, r, P, \gamma, \mu)$. $\mathcal{S}$ is the state space and $\mathcal{A}$ is the action space. The initial state $s_0$ is generated according to the distribution $\mu\in\Delta_\mathcal{S}$, where $\Delta_{\mathcal{S}}$ denotes the space of probability distributions over $\mathcal{S}$.
At timestep $t$, the agent observes from the environment a state $s_t\in\mathcal{S}$ and takes an action $a_t\in\mathcal{A}$.
Then the environment provides the agent with a reward signal $r(s_t,a_t)\in\mathbb{R}$. The state observation in the next timestep $s_{t+1}\in\mathcal{S}$ is sampled from the distribution $P(s_t, a_t)\in\Delta_\mathcal{S}$. We refer to $P\in(\Delta_\mathcal{S})^{\mathcal{S}\times\mathcal{A}}$ as the transition functions and $r:\mathbb{R}^{\mathcal{S}\times\mathcal{A}}$ as the reward function. A stationary stochastic policy $\pi\in(\Delta_\mathcal{A})^\mathcal{S}$ specifies a decision making strategy, where $\pi(s,a)$ is the probability of taking action $a$ in state $s$. The agent's goal is to learn an optimal policy $\pi^*$ that maximizes the expected cumulative reward:
\begin{equation}
    \pi^* = \argmax_{\pi\in\Pi} \mathbb{E}_{\pi,P} \sum_{t=0}^\infty \gamma^t r_t,
\end{equation}
where $\Pi$ denotes the policy space and $\gamma\in [0, 1)$ is the discount factor.

\subsection{Laplacian representation in RL}
\label{sec:bg-laprep}
The Laplacian Representation (\laprep)~\citep{wu2018laplacian} is a task-agnostic state representation in RL, originally proposed in~\cite{mahadevan2005proto} (known as proto-value function).
Formally, the Laplacian representations for all states are the eigenfunctions of Laplace-Beltrami diffusion operator on the state space manifold.
For simplicity, here we restrict the introduction of \laprep to the discrete state case and refer readers to~\citep{wu2018laplacian} for the formulation in the continuous case.

The states and transitions in an MDP can be viewed as nodes and edges in a graph.
The \laprep is formed by $d$ smallest eigenvectors of the graph Laplacian (usually $d\ll\lvert\mathcal{S}\rvert$). Each eigenvector (of length $\lvert\mathcal{S}\rvert$) corresponds to a dimension of the \laprep. Formally, we denote the graph as $\mathcal{G} = (\mathcal{S}, \mathcal{E})$ where $\mathcal{S}$ is the node set consisting of all states and $\mathcal{E}$ is the edge set consisting of transitions between states.
The Laplacian matrix of graph $\mathcal{G}$ is defined as $L\coloneqq D-A$, where $A$ is the adjacency matrix of $\mathcal{G}$ and $D\coloneqq\mathrm{diag}(A\mathbf{1})$ is the degree matrix~\citep{chung1997spectral}. 
We sort the eigenvalues of $L$ by their magnitudes and denote the $i$-th smallest one as $\lambda_i$. The unit eigenvector corresponding to $\lambda_i$ is denoted as $\mathbf{v}_i \in \mathbb{R}^{\lvert\mathcal{S}\rvert}$. 
Then, the $d$-dimensional \laprep of a state $s$ can be defined as
\begin{equation}
    \boldsymbol{\rho}_d (s)\coloneqq(\mathbf{v}_1[s], \mathbf{v}_2[s], \cdots, \mathbf{v}_d[s]),
\label{eqn:laprep}
\end{equation}
where $\mathbf{v}_i[s]$ denotes the entry in vector $\mathbf{v}_i$ corresponding to state $s$. 
In particular, $\mathbf{v}_1$ is a normalized all-ones vector and hence it provides no information about the environment geometry. Therefore we omit $\mathbf{v}_1$ and only consider other dimensions.

For environments with a large or even continuous state space, it is infeasible to obtain the \laprep by directly computing the eigendecomposition.
To approximate \laprep with neural networks, previous works~\cite{wu2018laplacian,wang2021towards} propose sample-based methods based on the spectral graph drawing~\cite{koren2005drawing}.
In particular, \citet{wang2021towards} introduce a generalized graph drawing objective that ensures dimension-wise faithful approximation to the ground truth $\boldsymbol{\rho}_d (s)$.

\section{Method}
\label{sec:method}
\subsection{Reachability-aware Laplacian representation}

\begin{figure}[t]
    \centering
    \includegraphics[width=\linewidth]{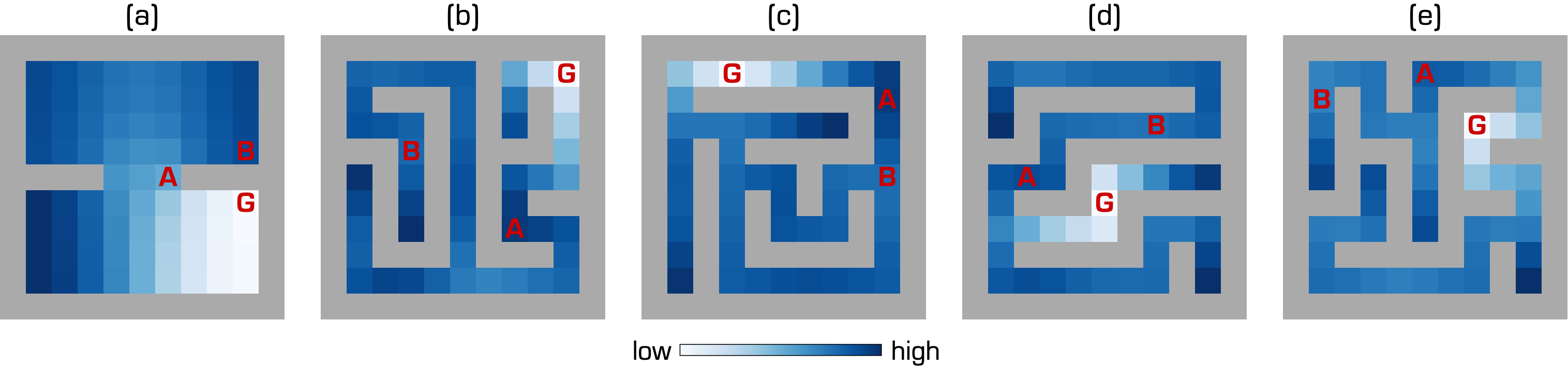}
    \caption{Visualizations of the Euclidean distances under \laprep between all states and the state \texttt{G}.}
     \label{fig:method}
\end{figure}

In prior works~\cite{wu2018laplacian, wang2021towards}, \laprep is believed to have a desirable property that the Euclidean distance between two states under \laprep (\ie, $\mathrm{dist}_{\boldsymbol{\rho}}(s,s')\coloneqq\lVert\boldsymbol{\rho}_d(s)-\boldsymbol{\rho}_d(s')\rVert$) roughly reflects the reachability between $s$ and $s'$.
That is, smaller distance between two states implies that it is easier for the agent to reach one state from the other.
Figure~\ref{fig:method}~(a) shows an illustrative example similar to the one in~\cite{wu2018laplacian}.
In this example, $\mathrm{dist}_{\boldsymbol{\rho}}(\texttt{A},\texttt{G})$ is smaller than $\mathrm{dist}_{\boldsymbol{\rho}}(\texttt{B},\texttt{G})$, which aligns with the intuition that moving to \texttt{G} from \texttt{A} takes fewer steps than from \texttt{B}.
Motivated by this, \laprep is used for reward shaping in goal-reaching tasks~\cite{wu2018laplacian, wang2021towards}.

However, little justification is provided in previous works~\cite{wu2018laplacian, wang2021towards} for this argument (\ie, the Euclidean distance under \laprep captures the inter-state reachability).
In fact, we find that it does not hold in general.
As shown in Figure~\ref{fig:method}~(b-e), $\mathrm{dist}_{\boldsymbol{\rho}}(\texttt{A},\texttt{G})$ is larger than $\mathrm{dist}_{\boldsymbol{\rho}}(\texttt{B},\texttt{G})$, but \texttt{A} is clearly closer to \texttt{G} than \texttt{B}.
As a result, when the agent moves towards the goal, $\mathrm{dist}_{\boldsymbol{\rho}}$ might give a wrong reward signal.
Such mismatch hinders the policy learning process when we use this distance for reward shaping.

In this paper, we introduce the following Reachability-Aware Laplacian Representation (\ralaprep):
\begin{equation}
    \boldsymbol{\phi}_d(s) \coloneqq \left(\frac{\mathbf{v}_2[s]}{\sqrt{\lambda_2}}, \frac{\mathbf{v}_3[s]}{\sqrt{\lambda_3}}, \cdots , \frac{\mathbf{v}_d[s]}{\sqrt{\lambda_d}}\right),
\label{eqn:ra-laprep}
\end{equation}
which can fix the issue of \laprep and better capture the reachability between states.
We provide both theoretical explanation (Section~\ref{sec:method-explain}) and empirical results (Section~\ref{sec:exp}) to demonstrate the advantage of \ralaprep over \laprep.

\subsection{Why RA-LapRep is more desirable than LapRep?}
\label{sec:method-explain}
In this subsection, we provide theoretical groundings for \ralaprep from two aspects, which explains why it better captures the inter-state reachability than \laprep.

First, we find that the Euclidean distance under \ralaprep is related to a quantity that measures the expected random walk steps between states.
Specifically, let $\mathrm{dist}_{\boldsymbol{\phi}}(s,s')\coloneqq\lVert\boldsymbol{\phi}_d(s)-\boldsymbol{\phi}_d(s')\rVert$ denote the Euclidean distance between states $s$ and $s'$ under \ralaprep.
When $d=\lvert\mathcal{S}\rvert$, $\mathrm{dist}_{\boldsymbol{\phi}}(s,s')$ has a nice interpretation~\cite{fouss2007random}: it is proportional to the square root of the average commute time between states $s$ and $s'$, \ie,
\begin{equation}
    \mathrm{dist}_{\boldsymbol{\phi}}(s,s') \propto \sqrt{n(s,s')}.
\label{eqn:proportional}
\end{equation}
Here the average commute time $n(s,s')$ measures the expected number of steps required in a random walk to navigate from $s$ to $s'$ and back (see Appendix for the formal definition).
Thus, $\mathrm{dist}_{\boldsymbol{\phi}}(s,s')$ provides a good quantification of the concept of reachability.
Additionally, with the proportionality in Eqn.~\eqref{eqn:proportional}, \ralaprep can be used to discover bottleneck states (see Section~\ref{sec:exp-bottleneck} for a detailed discussion and experiments).
In contrast, to the best of our knowledge, the Euclidean distance under \laprep (\ie,$\mathrm{dist}_{\boldsymbol{\rho}}(s,s')$) does not have a similar interpretation that matches the concept of reachability.

Second, we show that \ralaprep preserves global information while \laprep only focuses on preserving local information.
Specifically, we note that \ralaprep is equivalent (up to a constant factor) to the embedding obtained by classic Multi-Dimensional Scaling (MDS)~\cite{BorgGroenen2005} with the squared distance matrix in MDS being $D^{(2)}_{ij}=n(i,j)$~\cite{fouss2007random} (see Appendix for a detailed derivation).
Since classic MDS is known to preserve pairwise distances globally~\cite{tenenbaum2000global}, the Euclidean distance under \ralaprep is then a good fit for measuring the inter-state reachability.
In comparison, the \laprep is only able to preserve local information.
This is because, when viewing the MDP transition dynamic as a graph, the \laprep is essentially the Laplacian eigenmap~\cite{belkin2003laplacian}.
As discussed in \cite{belkin2003laplacian}, Laplacian eigenmap only aims to preserve local graph structure for each single neighborhood in the graph (\ie, mapping neighboring states close), while making no attempt in preserving global information about the whole graph (\eg, pairwise geodesic distances between nodes~\cite{tenenbaum2000global}).
Therefore, the Euclidean distance under \laprep is inherently not intended for measuring the reachability between states, particularly for distant states.

\subsection{Approximating RA-LapRep}
\label{sec:method-approx}
We note that the theoretical reasonings in above subsection are based on $d=\lvert\mathcal{S}\rvert$. 
In practice, however, for environments with a large or even continuous state space, it is infeasible to have $d=\lvert\mathcal{S}\rvert$ and hence we need to take a small $d$. 
One may argue that, using a small $d$ would lead to approximation error: when $d<\lvert\mathcal{S}\rvert$, the distance $\mathrm{dist}_{\boldsymbol{\phi}}(s,s')$ is not exactly proportional to $\sqrt{n(s,s')}$.
Fortunately, the gap between the approximated $\tilde{n}(s,s')$ and the true $n(s,s')$ turns out to be upper bounded by $C\sum_{i=d+1}^{\lvert\mathcal{S}\rvert}\frac{1}{\lambda_i}$, where $C$ is a constant and the summation is over the $\lvert\mathcal{S}\rvert-d$ largest eigenvalues.
Thus, this bound will not be very large.
We will empirically show in Section~\ref{sec:exp-shaping} that a small $d$ is sufficient for good reward shaping performance and further increasing $d$ does not yield any noticeable improvement.

Furthermore, even with a small $d$, it is still impractical to obtain \ralaprep via directly computing eigendecomposition.
To tackle this, we follow~\cite{wang2021towards} to approximate \ralaprep with neural networks using sample-based methods.
Specifically, we first learn a parameterized approximation $f_i(\cdot\,;\theta)$ for each eigenvector $\mathbf{v}_i$ by optimizing a generalized graph drawing objective~\cite{wang2021towards}, \ie, $f_i(s\,;\theta)\!\approx\!\mathbf{v}_i[s]$, where $\theta$ denotes the learnable parameters of the neural networks.
Next, we approximate each eigenvalue $\mathbf{\lambda}_i$ simply by
\begin{equation}
    \lambda_i = \mathbf{v}_i^\top L\mathbf{v}_i \approx \mathbb{E}_{(s,s')\in\mathcal{T}}\,\left(f_i(s\,;\hat{\theta}) - f_i(s'\,;\hat{\theta})\right)^2,
\end{equation}
where $\hat{\theta}$ denotes the learned parameters, and $\mathcal{T}$ is the same transition data used to train $f$.
Let $\tilde{\lambda}_i$ denote the approximated eigenvalue. \ralaprep can be approximated by
\begin{equation}
    \boldsymbol{\phi}_d(s) \approx \tilde{\boldsymbol{\phi}}_d(s) = \left(\frac{f_2(s\,;\hat{\theta})}{\sqrt{\tilde{\lambda}_2}}, \frac{f_3(s\,;\hat{\theta})}{\sqrt{\tilde{\lambda}_3}}, \cdots , \frac{f_d(s\,;\hat{\theta})}{\sqrt{\tilde{\lambda}_d}}\right).
\label{eqn:ra-laprep-approx}
\end{equation}
In experiments, we find this approximation works quite well and is on par with using the true $\boldsymbol{\phi}_d(s)$.

\section{Experiments}
\label{sec:exp}
\begin{figure}[t]
    \centering
    \includegraphics[width=0.95\linewidth]{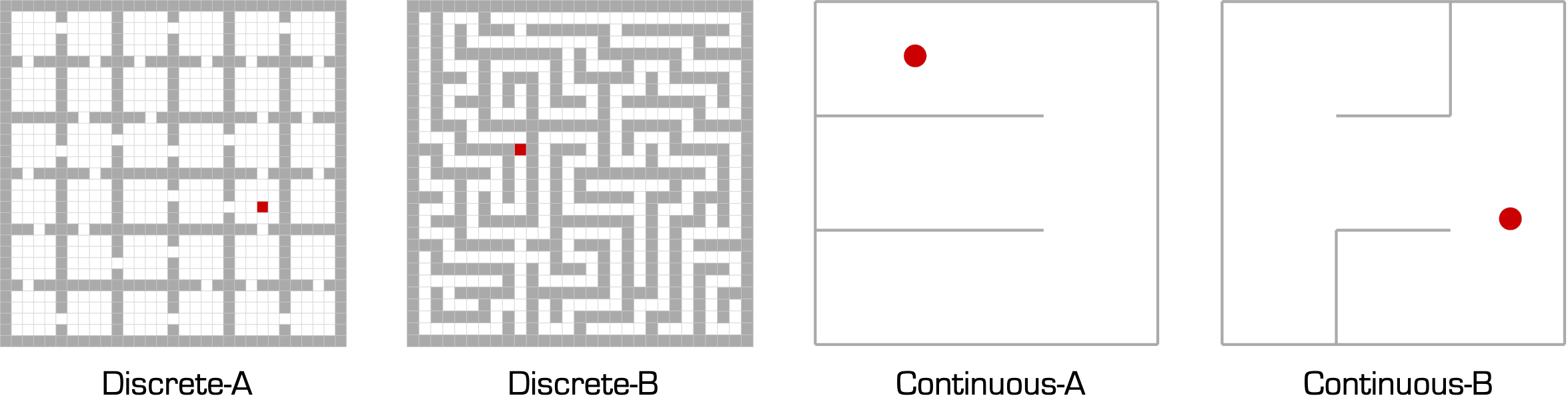}
    \caption{Environments used in our experiments (agents shown in red, and walls in grey).}
     \label{fig:environments}
\end{figure}

In this section, we conduct experiments to validate the benefits of \ralaprep compared to \laprep.
Following~\cite{wu2018laplacian, wang2021towards}, we consider both discrete gridworld and continuous control environments in our experiments. 
Figure~\ref{fig:environments} shows the layouts of environments used. We briefly introduce them here and refer readers to Appendix for more details. In discrete gridworld environments, the agent takes one of the four actions (\textit{up}, \textit{down}, \textit{left}, \textit{right}) to move from one cell to another.
If hitting the wall, the agent remains in the current cell. In continuous control environments, the agent picks a continuous action from $[-\pi,\pi)$ that specifies the direction along which the agent moves a fixed small step forward. For all environments, the observation is the agent's $(x,y)$-position.

\subsection{Capturing reachability between states}
\label{sec:exp-distance}

In this subsection, we evaluate the learned \ralaprep and \laprep in capturing the reachability among states.
We also include the ground-truth \ralaprep for comparison.
Specifically, for each state $s$, we compute the Euclidean distance between $s$ and an arbitrarily chosen goal state $s_{\textrm{goal}}$, under learned \laprep $\tilde{\boldsymbol{\rho}}$, learned \ralaprep $\tilde{\boldsymbol{\phi}}$, and ground-truth \ralaprep $\boldsymbol{\phi}$ (\ie, $\mathrm{dist}_{\tilde{\boldsymbol{\phi}}}(s,s_{\textrm{goal}})$, $\mathrm{dist}_{\tilde{\boldsymbol{\rho}}}(s,s_{\textrm{goal}})$ and $\mathrm{dist}_{\boldsymbol{\phi}}(s,s_{\textrm{goal}})$).
Then, we use heatmaps to visualize the three distances.
For continuous environments, the heatmaps are obtained by first sampling a set of states roughly covering the state space, and then performing interpolation among sampled states.

We train neural networks to learn \ralaprep and \laprep.
Specifically, we implement the two-step approximation procedure introduced in Section~\ref{sec:method-approx} to learn \ralaprep, and adopt the method in \cite{wang2021towards} to learn \laprep.
Details of training and network architectures are in Appendix.
The ground truth \ralaprep $\boldsymbol{\phi}$ is calculated using Eqn.~\eqref{eqn:ra-laprep}.
For discrete environments, the eigenvectors and eigenvalues are computed by eigendecomposition;
For continuous environments, the eigenfunctions and eigenvalues are approximated by the finite difference method with 5-point stencil~\cite{Knabner2003}.

\begin{figure}[t]
    \centering
    \includegraphics[width=\linewidth]{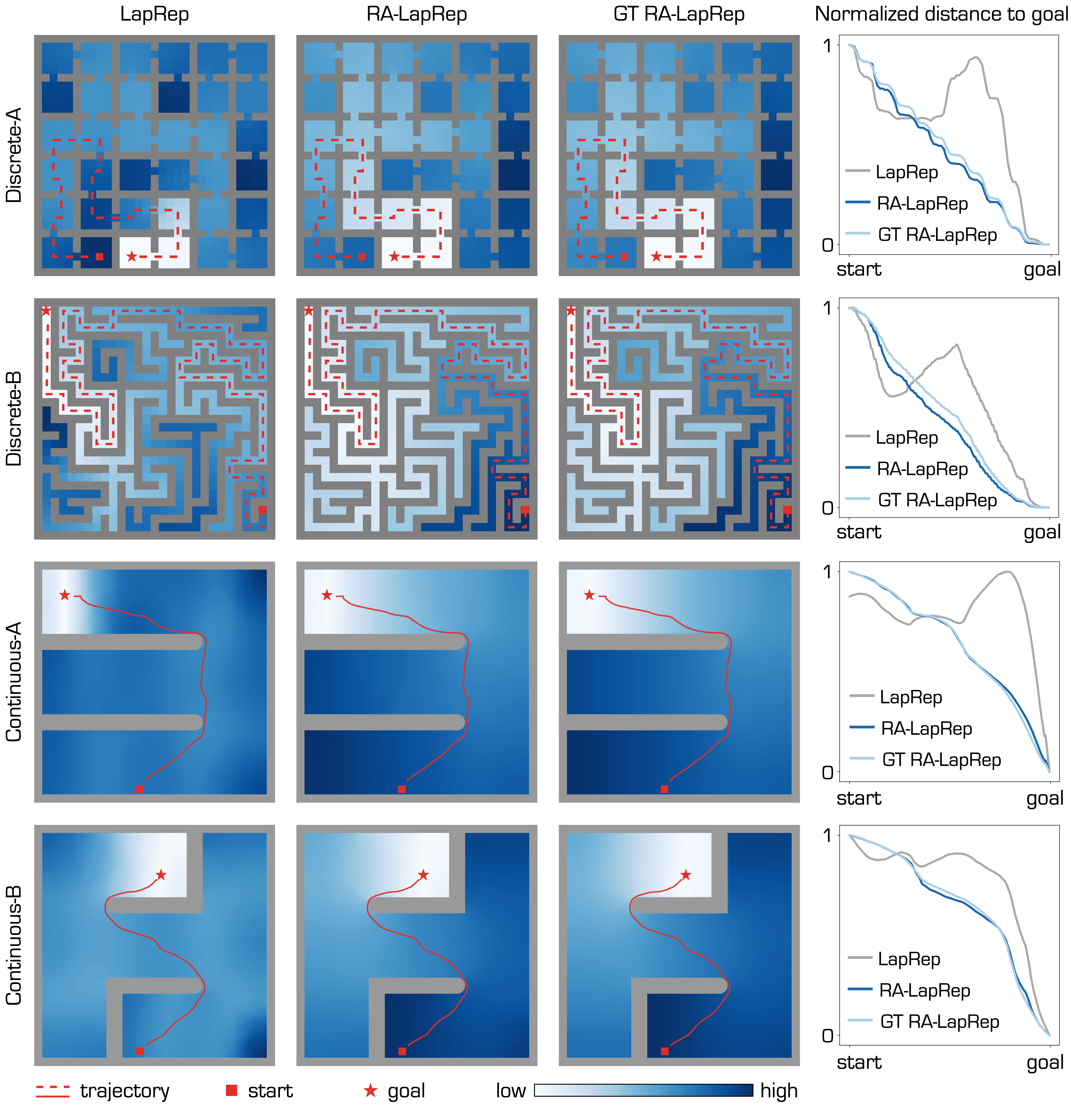}
    \caption{\textbf{Left 3 columns}: Visualizations of the Euclidean distances between all states and the goals in the four environments, under learned \laprep, learned \ralaprep, and ground-truth \ralaprep. Example trajectories are shown in red.
    \textbf{Right}: Line charts of the distance values for states in the trajectories (normalized to $[0, 1]$), where the states are sorted by temporal order.}
     \label{fig:dist}
\end{figure}

The visualization results are shown in Figure~\ref{fig:dist}.
For clearer comparison, we highlight in each environment an example trajectory, and plot the distance values along each trajectory in the line charts on the right.
As we can see, for both discrete and continuous environments, as the agent is moving towards the goal, the distances under the learned \laprep are increasing in some segments of the trajectories.
This is contradictory to that the Euclidean distances under \laprep reflects the inter-state reachability.
In contrast, the distances under \ralaprep decrease monotonically, which accurately reflect the reachability between current states and the goals.
We would like to note that, apart from the highlighted trajectories, similar observations can be obtained for other trajectories or goal positions (see Appendix).
Besides, we can see that the distances under the learned \ralaprep is very close to those under the ground-truth \ralaprep, indicating the effectiveness of our approximation approach.

\subsection{Reward shaping}
\label{sec:exp-shaping}
The above experiments show that the \ralaprep better captures the reachability between states than the \laprep.
Next, we study if this advantage leads to a higher performance for reward shaping in goal-reaching tasks.

Following \cite{wu2018laplacian, wang2021towards}, we define the shaped reward as
\begin{equation}
    r_t = 0.5 \cdot r_t^{\textrm{env}} + 0.5 \cdot r_t^{\textrm{dist}}.
\end{equation}
Here $r_t^{\textrm{env}}$ is the reward obtained from the environment, which is set to $0$ when the agent reaches the goal state and $-1$ otherwise.
For discrete environments, $r_t^{\textrm{env}}$ is simply formalized as $r_t^{\textrm{env}}=-\mathds{1}[s_{t+1}\ne s_\textrm{goal}]$.
For continuous environments, we consider the agent to have reached the goal when its distance to goal is within a small preset radius $\epsilon$, \ie, $r_t^{\textrm{env}}=-\mathds{1}[\lVert s_{t+1} - s_\textrm{goal} \rVert > \epsilon]$.
The pseudo-reward $r_t^{\textrm{dist}}$ is set to be the negative distance under the learned representations:
\begin{equation}
    \begin{aligned}
         r_t^{\textrm{dist}}=-\mathrm{dist}_{\tilde{\boldsymbol{\rho}}}(s_{t+1},s_\textrm{goal}) & \quad \text{for \laprep}, \\
         r_t^{\textrm{dist}}=-\mathrm{dist}_{\tilde{\boldsymbol{\phi}}}(s_{t+1},s_\textrm{goal}) & \quad \text{for \ralaprep}.
    \end{aligned}
\end{equation}
As in \cite{wu2018laplacian, wang2021towards}, we also include two baselines: $L_2$ shaping, \ie, $r_t^{\textrm{dist}}=-\lVert s_{t+1}-s_\textrm{goal}\rVert$, and no reward shaping, \ie, $r_t = r_t^{\textrm{env}}$.
Following~\cite{wang2021towards}, we consider multiple goal positions for each environment (see Appendix), in order to minimize the bias brought by goal positions.
The final results are averaged across different goals and 10 runs per goal.

\begin{figure}[t]
    \centering
    \includegraphics[width=\linewidth]{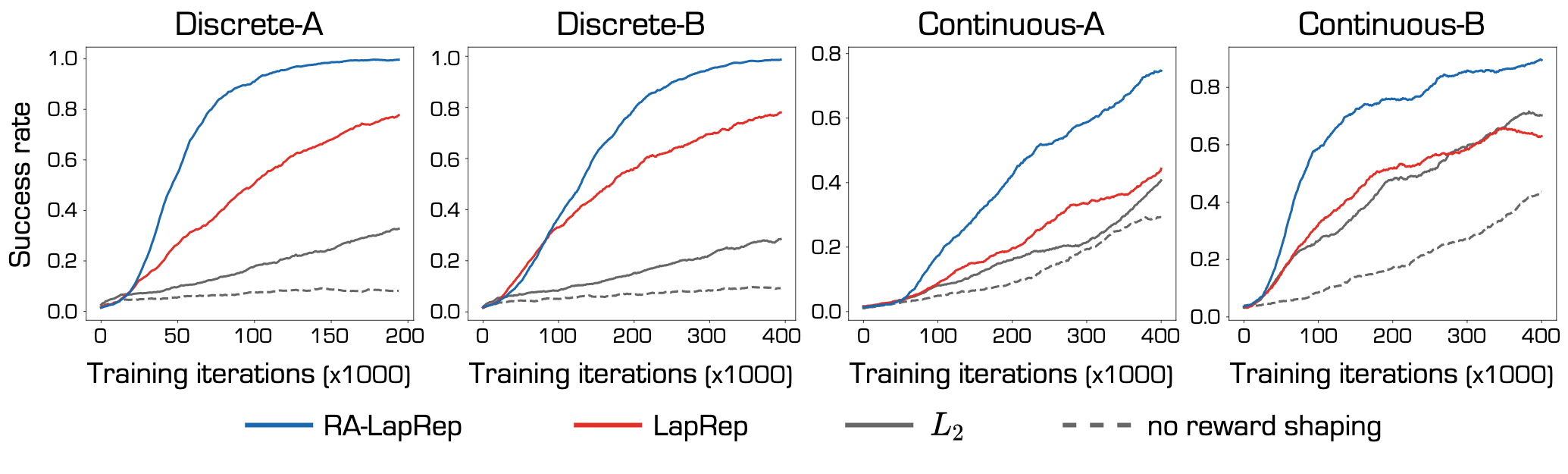}
    \caption{Reward shaping results in goal-reaching tasks, with different choices of the shaped reward.}
     \label{fig:reward_shaping}
\end{figure}

As shown in Figure~\ref{fig:reward_shaping}, on both discrete and continuous environments, \ralaprep outperforms \laprep and other two baselines by a large margin.
Compared to \laprep, using \ralaprep for reward shaping is more sample efficient, reaching the same level of performance in fewer than half of steps.
We attribute this performance improvement to that the Euclidean distance under \ralaprep more accurately captures the inter-state reachability.

\begin{figure}[t]
    \centering
    \includegraphics[width=\linewidth]{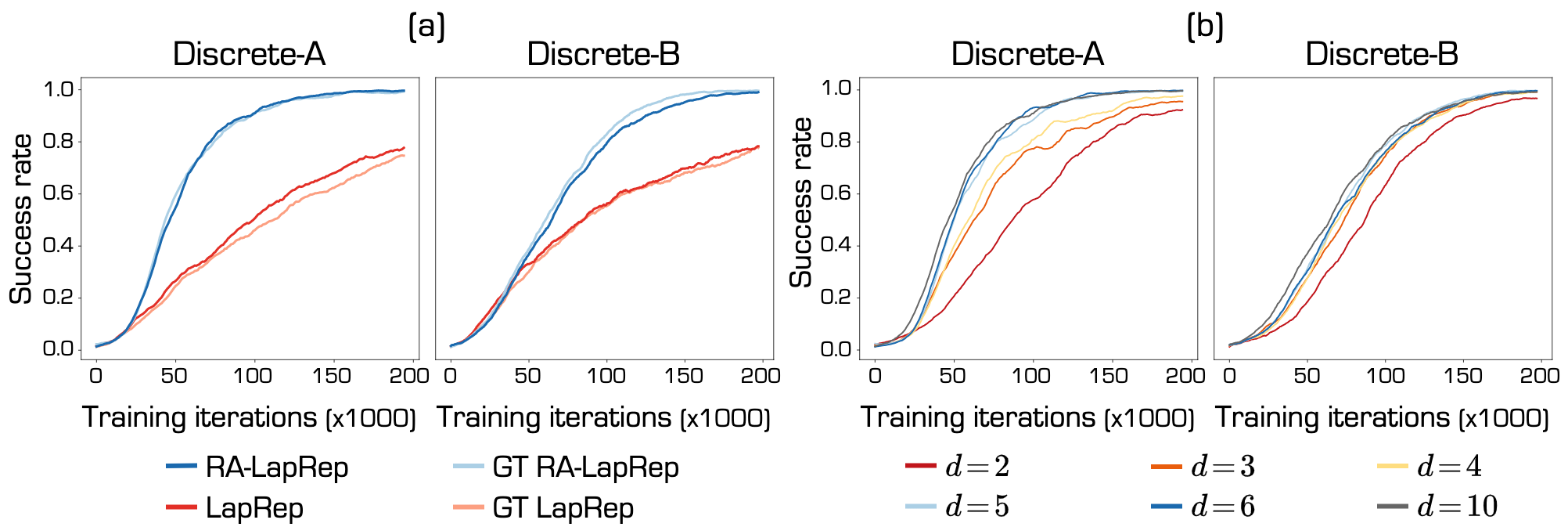}
    \caption{
    \textbf{(a)} Comparison in the reward shaping results among the learned representations and the ground-truth ones; \textbf{(b)} Comparison in the reward shaping results among using different $d$ for the learned \ralaprep.}
     \label{fig:ablation}
\end{figure}

\paragraph{Comparing to the ground-truth}

To find out if the neural network approximation limits the performance, we also use the ground-truth \ralaprep and \laprep for reward shaping, and compare the results with those of the learned representations.
From Figure~\ref{fig:ablation}, we can see that the performance of the learned \ralaprep is as good as the ground-truth one, indicating the effectiveness of the network approximation.
Besides, the performances of the learned \laprep and the ground-truth one are very close, suggesting that the inferior performance of \laprep is not due to poor approximation.

\paragraph{Varying the dimension $d$}
As mentioned in Section~\ref{sec:method-approx}, the theoretical approximation error of using a smaller $d$ is not very large. Here we conduct experiments to see if this error cause significant performance degradation. Specifically, we vary the dimension $d$ from 2 to 10, and compare the resulting reward shaping performance. As Figure~\ref{fig:ablation} shows, the performance first improves as we increase $d$, and then plateaus. Thus, a small $d$ (\eg, $d=10$, compared to $\lvert\mathcal{S}\rvert>300$) is sufficient to give a pretty good performance.

\subsection{Discovering bottleneck states}
\label{sec:exp-bottleneck}
\begin{figure}[t]
    \centering
    \includegraphics[width=\linewidth]{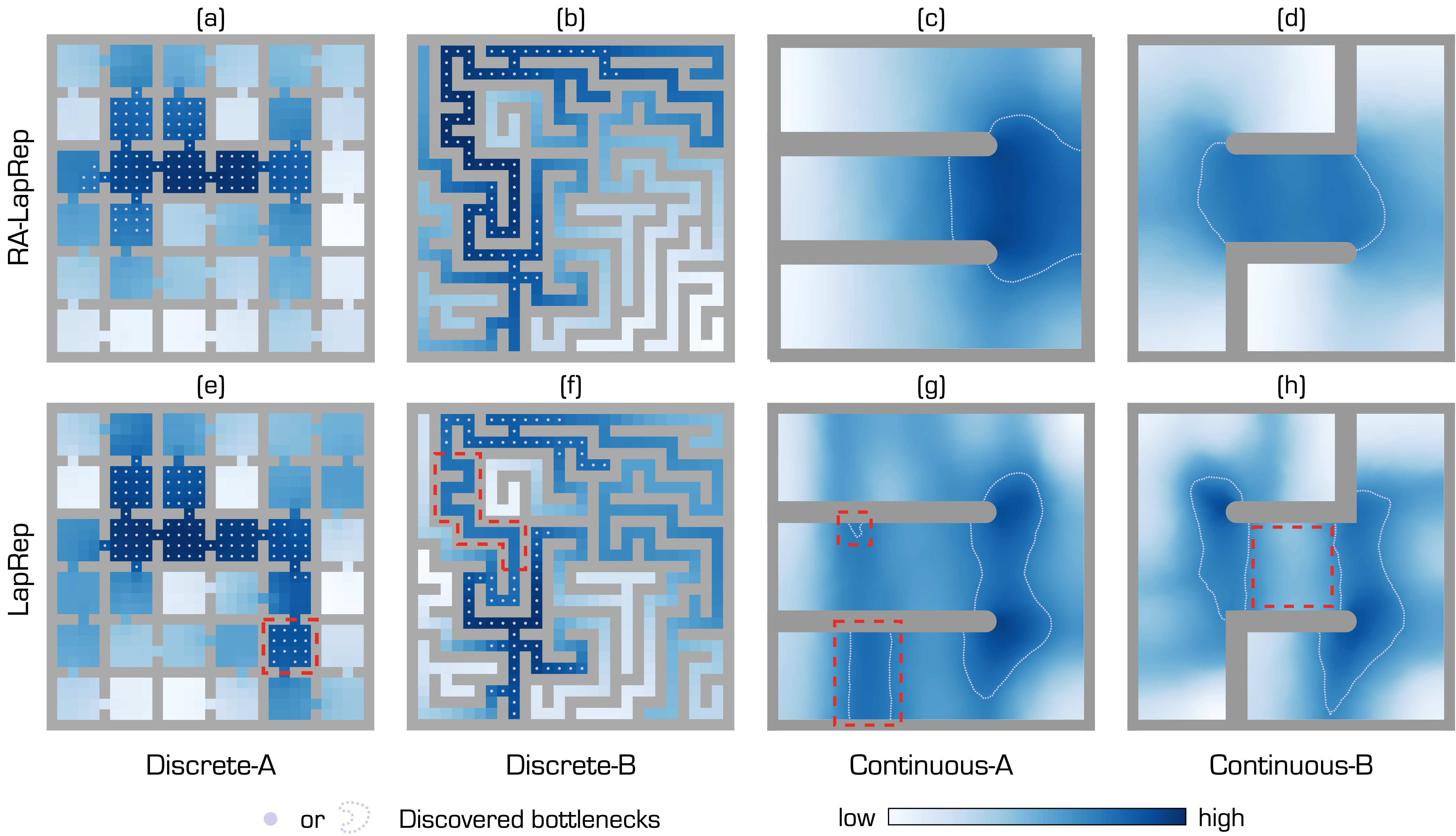}
    \caption{Bottleneck discovery results of the learned \ralaprep and \laprep. Discovered bottleneck states are those marked as dots (for discrete environments), or those within the contour lines (for continuous environments).}
     \label{fig:bottleneck}
\end{figure}

The bottleneck states are analogous to the key nodes in graph, which allows us to discover them based on the graph centrality measure.
Here we consider a simple definition of the centrality~\cite{bavelas1950communication}:
\begin{equation}
    \mathrm{cent}(s)=\left(\sum_{s'\in\mathcal{S}} \mathrm{dist}(s, s')\right)^{-1},
\label{eqn:centrality}
\end{equation}
where $\mathrm{dist}$ is a generic distance measure.
The states with high centrality are those where many paths pass, hence they can be considered as bottlenecks.
Since the Euclidean distance under \ralaprep more accurately reflects the inter-state reachability, we aim to see if it benefits bottleneck discovery.

Specifically, we calculate $\mathrm{cent}(s)$ for all states with $\mathrm{dist}$ being the Euclidean distance under the learned \ralaprep or the learned \laprep, and take top $20\%$ states with lowest $\mathrm{cent}(s)$ as the discovered bottlenecks.
For continuous environments, the summation in Eqn.~\eqref{eqn:centrality} is calculated over a set of sampled states.
Figure~\ref{fig:bottleneck} visualizes the computed $\mathrm{cent}(s)$ and highlights the discovered bottleneck states.
In this top row of Figure~\ref{fig:bottleneck}, we can see that the bottlenecks discovered by the learned \ralaprep are essentially the states that many paths pass through.
In comparison, the results of using \laprep are not as satisfactory.
For one thing, as highlighted (with dashed red box) in Figure~\ref{fig:bottleneck}~(e) and (g), some of the discovered states are actually not in the center of the environment (\ie, where most trajectories pass), which does not match the concept of bottleneck.
For another, as highlighted in Figure~\ref{fig:bottleneck}~(f) and (h), some regions that should have been identified are however missing.

\section{Discussions}
\label{sec:discuss}

\subsection*{Predicting the average commute time directly}

One may wonder if it is possible to directly predict the average commute time (\ie, in a supervised learning fashion), in contrast to our training approach (which can be viewed an unsupervised way). For example, we can minimize the following error to train the representation
\begin{equation}
    \mathbb{E}_{s_i,s_j} \left[\lVert \phi(s_i) - \phi(s_j) \rVert - n(s_i, s_j)\right],
\end{equation}
where $\phi(s)$ denote the state representation. However, obtaining accurate $n(s_i,s_j)$ requires the knowledge of the whole graph, which is infeasible in general. One workaround is to sample $s_i$ and $s_j$ from the same trajectory and use the difference between their temporal indices as a proxy of $n(s_i,s_j)$. But such approximation suffers from high variance. With poor estimation of $n(s_i,s_j)$, the learned representation will consequently be of low quality. In comparison, our RA-LapRep can be viewed as an unsupervised learning method, which does not reply on estimating $n(s_i,s_j)$.

\subsection*{Generalization to directed graphs}

One implicit assumption in our and previous works~\cite{wu2018laplacian, wang2021towards} is that the underlying graph is undirected, which implies that the actions are reversible. In practical settings such as robot control, this is often not the case. To tackle this limitation, one way is generalizing the notion of \ralaprep to directed graphs, for example, by taking inspiration from effective resistance on directed graphs~\cite{young2016new,Boley2011commute, fitch2019effective}. 

However, it is a highly non-trivial challenge. First, it is not straightforward to give a proper definition for \ralaprep in directed cases. Moreover, due to the complex-valued eigenvalues of directed graph Laplacian matrices, designing an optimization objective to approximate the eigenvectors (as done in \cite{wu2018laplacian, wang2021towards}) would be difficult. Despite the challenges, generalization to directed graphs would be an interesting research topic and worth an in-depth study beyond this work.

\subsection*{Evaluation on high-dimensional environments}

In this work, we use 2D mazes because such environments allow us to easily examine whether the inter-state reachability is well captured. For applying our method to more complex environments such as Atari~\cite{bellemare2013arcade}, we foresee some non-trivial challenges. For example, most games contain irreversible transitions. So when dealing with these environments, we may need to first generalize our method to directed graphs. Nevertheless, as a first step towards applying \ralaprep to high-dimensional environments, we conduct experiments by replacing the 2D $(x,y)$ position input with the high-dimensional top-view images input. The experiment results (see Appendix) show that, with high-dimensional input, the learned \ralaprep is still able to
accurately reflect the reachability among states and significantly boost the reward shaping performance. This suggests that learning \ralaprep with high-dimensional input on more complex environments is possible, and we will continue to explore along this direction in future works.

\subsection*{Ablation on uniform state coverage}

Follow \cite{wu2018laplacian, wang2021towards}, we train \ralaprep with pre-collected transition data that roughly uniformly covers the state space.
One may wonder how the learned \ralaprep would be affected when the uniformly full state coverage assumption breaks.
To investigate this, we conduct ablative experiments (learning \ralaprep and reward shaping) by manipulating the distribution of collected data.
The results show that, the learned \ralaprep is robust to moderate changes in the data distribution, \wrt both its capacity in capturing the reachability and its reward shaping performance. Only when the distribution is too non-uniform, the resulted graph will be disconnected and the performance will degrade.
Please refer to the Appendix for details about the experiments setup and results.

\section{Related Works}
\label{sec:related}

Our work is built upon prior works on learning Laplacian representation with neural networks~\cite{wu2018laplacian, wang2021towards}. We introduce \ralaprep that can more accurately reflect the inter-state reachability. Apart from Laplacian representations, another line of works also aims to learn representation that captures the inter-state reachability~\cite{hartikainen2020dynamical,savinov2018episodic,zhang2020generating}.
However, their learned reachability is not satisfying (see Appendix for detailed discussions).

Apart from reward shaping, Laplacian representations have also found applications in option discovery~\cite{machado2017laplacian, machado2018eigenoption, jinnai2019exploration, wang2021towards}.
We would like to note that our \ralaprep can still be used for option discovery and would yield same good results as \laprep~\cite{machado2017laplacian, wang2021towards}, since the dimension-wise scaling (in Eqn.~\ref{eqn:ra-laprep}) does not change the eigen-options.
Regarding bottleneck state discovery, there are prior works~\cite{simsek2008skill, moradi2010automatic} that adopt other centrality measures in graph theory (\eg, betweenness) to find bottleneck states for skill characterization.

\section{Conclusion}
\label{sec:conclusion}

Laplacian Representation (\laprep) is a task-agnostic state representation that encodes the geometry structure of the environment. In this work, we point out a misconception in prior works that the Euclidean distance in the \laprep space can reflect the reachability among states. We show that this property does not actually hold in general, \ie, two distant states in the environment may have small distance under \laprep. Such issue would limit the performance of using this distance for reward shaping~\cite{wu2018laplacian, wang2021towards}.

To fix this issue, we introduce a Reachability-Aware Laplacian Representation (\ralaprep). Compared to \laprep, we show that the Euclidean distance \ralaprep provides a better quantification of the inter-state reachability. Furthermore, this advantage of \ralaprep leads to significant performance improvements in reward shaping experiments. In addition, we also provide theoretical explanation for the advantages of \ralaprep.

\putbib
\end{bibunit}

\begin{bibunit}
\clearpage
\appendix

\section{Definitions and Derivations}

\begin{definition}[\textit{Average First-Passage Time}~\cite{fouss2007random,kemeny1983finite}]
Given a finite and connected graph $\mathcal{G}$, the \textit{average first-passage time} $m(j|i)$ is defined as the average number of steps required in a random walk that starts from node $i$, to reach node $j$ for the first time.
\end{definition}

\begin{definition}[\textit{Average Commute Time}~\cite{fouss2007random,gobel1974random}]
Given a finite and connected graph $\mathcal{G}$, the \textit{average commute time} $n(i,j)$ is defined as the average number of steps required in a random walk that starts from node $i$, to reach node $j$ for the first time and then go back to
node $i$. That is, $n(i,j)=m(j|i)+m(i|j)$.
\end{definition}

\subsection{Connection between RA-LapRep and the average commute time}

Let us first set up some notations. We denote the pseudo-inverse of the graph Laplacian matrix $L$ as $L^+$.
We denote the $i$-th smallest eigenvalue (sorted by magnitude) of $L^+$ as $\lambda^+_i$, and the corresponding unit eigenvector as $\mathbf{u}_i$.
Recall that we denote the $i$-th smallest eigenvalue (sorted by magnitude) of $L$ as $\lambda_i$, and the corresponding unit eigenvector as $\mathbf{v}_i$.
Assuming the graph is connected, we have the following correspondence between the eigenvectors/eigenvalues of $L^+$ and those of $L$:
\begin{equation}
\begin{aligned}
    \mathbf{u}_1 &= \mathbf{v}_1, \\
    (\mathbf{u}_2, \mathbf{u}_3, \cdots, \mathbf{u}_{\lvert\mathcal{S}\rvert}), &= (\mathbf{v}_{\lvert\mathcal{S}\rvert}, \mathbf{v}_{\lvert\mathcal{S}\rvert-1}, \cdots, \mathbf{v}_2) \\
    \lambda^+_1 &= \lambda_1, \\
    (\lambda^+_2, \lambda^+_3, \cdots, \lambda^+_{\lvert\mathcal{S}\rvert-1}) &= ( \frac{1}{\lambda_{\lvert\mathcal{S}\rvert}}, \frac{1}{\lambda_{\lvert\mathcal{S}\rvert-1}}, \cdots, \frac{1}{\lambda_2}).
\label{eqn:pseudo-inv}
\end{aligned}
\end{equation}
In particular, $\mathbf{u}_1=\mathbf{v}_1$ is a normalized all-ones vector and $\lambda^+_1=\lambda_1=0$.

Let $U=(\mathbf{u}_1, \mathbf{u}_2, \cdots, \mathbf{u}_{\lvert\mathcal{S}\rvert})$ and $\Lambda=\mathrm{diag}(\lambda^+_1, \lambda^+_2, \cdots, \lambda^+_{\lvert\mathcal{S}\rvert})$.
We use $\mathbf{e}_s\in\mathbb{R}^\mathcal{S}$ to denote a standard unit vector with $i$-th entry being 1. 
\citet{fouss2007random} show the connection between the average commute time and the eigenvectors of the pseudo-inverse of Laplacian matrix $L^+$,
\begin{equation}
    n(s,s')= V_\mathcal{G} \lVert \Lambda^{\frac{1}{2}} U^\top\mathbf{e}_s - \Lambda^{\frac{1}{2}} U^\top\mathbf{e}_{s'} \rVert ^2.
\label{eqn:commute-pesudo-inv}
\end{equation}
where $V_\mathcal{G}$ is the volume of graph $\mathcal{G}$ (\ie, sum of the node degrees).
Since $\lambda^+_1=0$, we can obtain
\begin{equation}
    n(s,s')= V_\mathcal{G} \left\lVert \left(\sqrt{\lambda^+_2}\mathbf{u}_2, \cdots, \sqrt{\lambda^+_{\lvert\mathcal{S}\rvert}}\mathbf{u}_{\lvert\mathcal{S}\rvert}\right)^\top\mathbf{e}_s - \left(\sqrt{\lambda^+_2}\mathbf{u}_2, \cdots, \sqrt{\lambda^+_{\lvert\mathcal{S}\rvert}}\mathbf{u}_{\lvert\mathcal{S}\rvert}\right)^\top\mathbf{e}_{s'} \right\rVert ^2.
\end{equation}
Based on Eqn~\eqref{eqn:pseudo-inv}, we can get
\begin{equation}
    n(s,s') = V_\mathcal{G} \left\lVert \left(\frac{\mathbf{v}_{\lvert\mathcal{S}\rvert}}{\sqrt{\lambda_{\lvert\mathcal{S}\rvert}}}, \cdots, \frac{\mathbf{v}_2}{\sqrt{\lambda_2}}\right)^\top\mathbf{e}_{s} - \left(\frac{\mathbf{v}_{\lvert\mathcal{S}\rvert}}{\sqrt{\lambda_{\lvert\mathcal{S}\rvert}}}, \cdots, \frac{\mathbf{v}_2}{\sqrt{\lambda_2}}\right)^\top\mathbf{e}_{s'} \right\rVert ^2.
\end{equation}
Therefore, when $d=\lvert\mathcal{S}\rvert$, we have
\begin{equation}
    n(s,s')=V_\mathcal{G}\lVert \boldsymbol{\phi}_{d}(s) - \boldsymbol{\phi}_{d}(s') \rVert ^2 = V_\mathcal{G}\, (\mathrm{dist}_{\boldsymbol{\phi}}(s,s'))^2,
\end{equation}
and that is $\mathrm{dist}_{\boldsymbol{\phi}}(s,s')\propto\sqrt{n(s,s')}$.

\subsection{Equivalence between RA-LapRep and MDS}

Given an input matrix of pairwise dissimilarities between $n$ items, classic MDS~\cite{BorgGroenen2005} outputs an embedding for each item, such that the pairwise Euclidean distances between embeddings preserve the pairwise dissimilarities as well as possible.

Specifically, given the squared dissimilarity matrix $D^{(2)}\in\mathbb{R}^{n\times n}$, MDS first applies doubly centering on $D^{(2)}$:
\begin{equation}
    B = -\frac{1}{2}JD^{(2)}J,
\end{equation}
where $J=I-\frac{1}{n}\mathbf{1}\mathbf{1}^\top$ is a centering matrix. Next, MDS calculates the eigendecomposition of $B$. Let $\Lambda_+$ denote the diagonal matrix containing the eigenvalues greater than 0, and $Q_+$ denote the matrix made of corresponding eigenvectors. Then the embedding matrix is given by $X=Q_+\Lambda_+^{\frac{1}{2}}$.

Let $N$ denote the matrix containing pairwise average commute time, \ie, $[N]_{ij}=n(i,j)$. Then we have
\begin{equation}
    -\frac{1}{2}[JNJ]_{ij} = -\frac{1}{2}\left[n(i,j) - \frac{1}{\lvert\mathcal{S}\rvert}\sum_{k=1}^{\lvert\mathcal{S}\rvert}n(i,k) - \frac{1}{\lvert\mathcal{S}\rvert}\sum_{h=1}^{\lvert\mathcal{S}\rvert}n(h,j) + \frac{1}{\lvert\mathcal{S}\rvert^2}\sum_{h=1}^{\lvert\mathcal{S}\rvert}\sum_{k=1}^{\lvert\mathcal{S}\rvert}n(h,k) \right].
\end{equation}
\citet{fouss2007random} show that
\begin{equation}
    n(i,j)=V_\mathcal{G}(l^+_{ii} + l^+_{jj} - 2l^+_{ij})
\end{equation}
where $l^+_{ij}=[L^+]_{ij}$. Thus, we can get
\begin{equation}
\begin{aligned}
    -\frac{1}{2}[JNJ]_{ij}
    &= -\frac{1}{2}V_\mathcal{G} \left[ - 2l^+_{ij} + \frac{1}{\lvert\mathcal{S}\rvert}\sum_{k=1}^{\lvert\mathcal{S}\rvert} 2l^+_{ik} + \frac{1}{\lvert\mathcal{S}\rvert}\sum_{h=1}^{\lvert\mathcal{S}\rvert} 2l^+_{hj} - \frac{1}{\lvert\mathcal{S}\rvert^2}\sum_{h=1}^{\lvert\mathcal{S}\rvert}\sum_{k=1}^{\lvert\mathcal{S}\rvert} 2l^+_{hk}
    \right] \\
    &= V_\mathcal{G} \left[ l^+_{ij} - \frac{1}{\lvert\mathcal{S}\rvert}\sum_{k=1}^{\lvert\mathcal{S}\rvert} l^+_{ik} - \frac{1}{\lvert\mathcal{S}\rvert}\sum_{h=1}^{\lvert\mathcal{S}\rvert} l^+_{hj} + \frac{1}{\lvert\mathcal{S}\rvert^2}\sum_{h=1}^{\lvert\mathcal{S}\rvert}\sum_{k=1}^{\lvert\mathcal{S}\rvert} l^+_{hk}
    \right].
\end{aligned}
\end{equation}
Since $L^+$ is doubly centered, \ie, the sum of each row or each column is 0 (see Appendix A.1.2 in~\cite{fouss2007random}), we can obtain
\begin{equation}
    -\frac{1}{2}[JNJ]_{ij} = V_\mathcal{G} l^+_{ij},
\end{equation}
that is, $B=V_\mathcal{G}L^+$ with $D^{(2)}=N$.
Therefore, the embeddings from MDS is equivalent to the embeddings $U\Lambda^\frac{1}{2}$ in Eqn.~\eqref{eqn:commute-pesudo-inv} (up to a constant $\sqrt{V_\mathcal{G}}$). Based on the derivations in previous subsection, we can see the equivalence between \ralaprep and the embeddings from MDS.

\section{Environment descriptions}

The two discrete environments used in our experiments are built with MiniGrid~\cite{gym_minigrid}.
Specifically, the \texttt{Discrete-A} environment is a $29\times29$ grid with 391 states, and the \texttt{Discrete-B} environment is a $31\times31$ grid with 611 states.
The agent has 4 four actions: \textit{moving left}, \textit{moving right}, \textit{moving up} and \textit{moving down}.
We consider two kinds of raw state representations : $(x, y)$ position and top-view image of the grid.
To pre-process the input for network training, we scale $(x, y)$ positions to the range $[-0.5, 0.5]$, and the top-view image to the range $[0, 1]$.

The two continuous environments used in our experiments are built with MuJoCo~\cite{todorov2012mujoco}.
Specifically, both \texttt{Continuous-A} and \texttt{Continuous-B} are of size $15\times15$.
A ball with diameter 1 is controlled to navigate in the environment.
At each step, the agent takes a continuous action (within range $[-\pi,\pi)$) that specifies a direction, and then move a small step forward along this direction (step size set to 0.1).
We consider the $(x, y)$ positions as the raw state representations, and also scale them to the range $[-0.5,0.5]$ in pre-processing.

\section{Experiment details}

\subsection{Learning the representations (LapRep and RA-LapRep)}

To learn \laprep, we follow the training setup in~\cite{wang2021towards} (see their Appendix~E.1).
Briefly speaking, we first collect a dataset of transitions using a uniformly random policy with random starts, and then learn \laprep on this dataset with mini-batch gradient decent. We use the same network architectures as in~\cite{wang2021towards} and the Adam optimizer~\cite{kingma2014adam}. Other configurations are summarized in Table~\ref{tab:configs-laprep}. After learning \laprep, we approximate \ralaprep with the approach introduced in Section~\ref{sec:method-approx}.

\begin{table}[h]
    \centering
    \caption{Configurations for learning \laprep.}
    \begin{tabular}{ccc}
    \toprule
    & Discrete environments & Continuous environments \\
    \midrule
    Dataset size (in terms of steps) & 100,000 & 1,000,000 \\
    Episode length & 50 & 500 \\
    Training iterations & 200,000 & 400,000\\
    Learning rate & 1e-3 & 1e-3 \\
    Batch size & 1024 & 8192 \\
    \laprep dimension $d$ & 10 & 10 \\
    Discount sampling & 0.9 & 0.95 \\
    \bottomrule
    \end{tabular}
    \label{tab:configs-laprep}
\end{table}

\subsection{Reward shaping}

\begin{figure}[t]
    \centering
    \includegraphics[width=\linewidth]{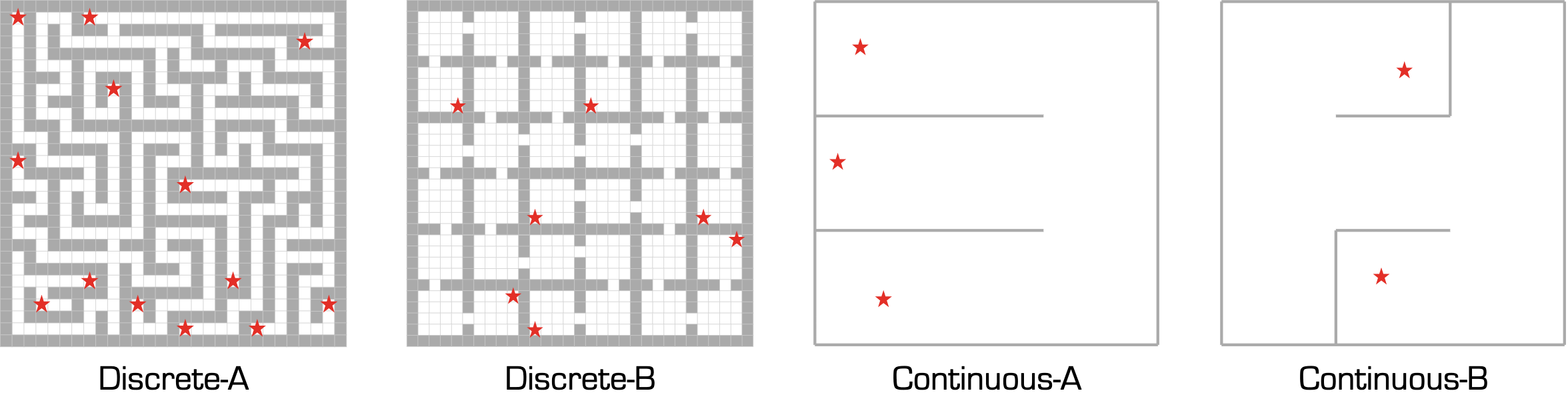}
    \caption{Goal positions for reward shaping experiments. Each red star represents a goal.}
     \label{fig: goal-positions}
\end{figure}

Following~\cite{wu2018laplacian, wang2021towards}, we train the agent in goal-achieving tasks using Deep Q-Network (DQN)~\cite{mnih2013playing} for discrete environments, and Deep Deterministic Policy Gradient (DDPG)~\cite{lillicrap2016continuous} for continuous environments.
We use the $(x,y)$ position observation in both discrete and continuous cases.
For DQN, we use the same fully-connected network as in~\cite{wang2021towards}.
For DDPG, we use a two-layer fully connected neural network with units $(400, 300)$ for both the actor and the critic.
The detailed configurations are summarized in Table~\ref{tab:configs-rew-shaping}. As mentioned in Section~\ref{sec:exp-shaping}, we consider multiple goals to minimize the bias brought by the goal positions. The locations of these goals are shown in Figure~\ref{fig: goal-positions}.

\begin{table}[h]
    \centering
    \caption{Configurations for reward shaping.}
    \begin{tabular}[t]{cc}
    \toprule
    \multicolumn{2}{c}{DQN} \\
    \midrule
    \multirow{2}{*}{Timesteps} & 200,000 (\texttt{Discrete-A}) \\
    & 400,000 (\texttt{Discrete-B}) \\
    Episode length & 150 \\
    Optimizer & Adam \\
    Learning rate & 1e-3 \\
    Learning starts & 5,000 \\
    Training frequency & 1 \\
    Target update frequency & 50 \\
    Target update rate & 0.05 \\
    Replay size & 100,000 \\
    Batch size & 128 \\
    Discount factor $\gamma$ & 0.99 \\
    \bottomrule
    \end{tabular}
    \quad
    \begin{tabular}[t]{cc}
    \toprule
    \multicolumn{2}{c}{DDPG} \\
    \midrule
    Timesteps & 500,000 \\
    Episode length & 1,000 \\
    Optimizer & Adam \\
    Learning rate & 1e-3 \\
    Learning starts & 100,000 \\
    Target update rate & 0.001 \\
    Replay size & 200,000 \\
    Batch size & 2,048 \\
    Discount factor $\gamma$ & 0.99 \\
    Action noise type & Gaussian noise \\
    Gaussian noise $\sigma$ & 0.5 \\
    \bottomrule
    \end{tabular}
    \label{tab:configs-rew-shaping}
\end{table}

\subsection{Resource types and computation cost}

Our experiments are run on Linux servers with Intel\textsuperscript{\textregistered} Core\textsuperscript{TM} i7-5820K CPU and NVIDIA Titan X GPU.
For discrete environments with $(x,y)$ positions as observations, each run of representation learning requires about 600MB GPU memory and takes about 50 minutes.
Each run of reward shaping requires about 600MB GPU memory and takes about 20 minutes (for \texttt{Discrete-A}) or 40 minutes (for \texttt{Discrete-B}).
For discrete environments with top-view images as observations, each run of representation learning requires about 900MB GPU memory and takes about 1 hour.
For continuous environments, each run of representation learning requires about 4GB GPU memory and takes about 11 hours. Each run of reward shaping requires about 900MB GPU memory and takes about 1.5 hours.

\section{Additional results}

\subsection{Capturing reachability between states}

\begin{figure}[h]
    \centering
    \includegraphics[width=\linewidth]{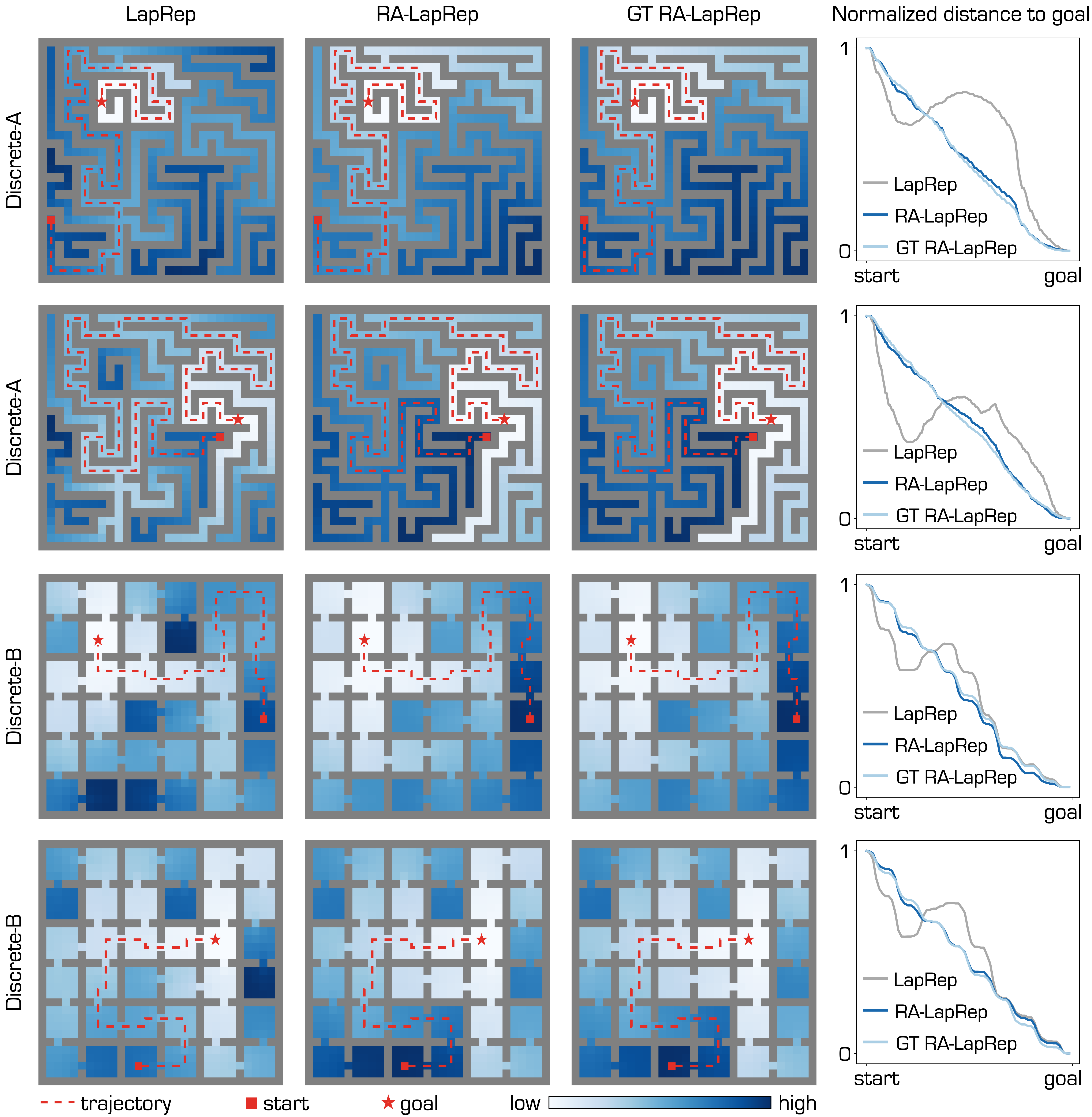}
    \caption{\textbf{Left 3 columns}: Visualization of the Euclidean distance between all states and the goals in discrete environments. For each environment, two additional trajectories (different from the one in the main paper) are shown in red. \textbf{Right}: Normalized distance values for states in the trajectories.}
     \label{fig: supp-dist_discrete}
\end{figure}

\begin{figure}[h]
    \centering
    \includegraphics[width=\linewidth]{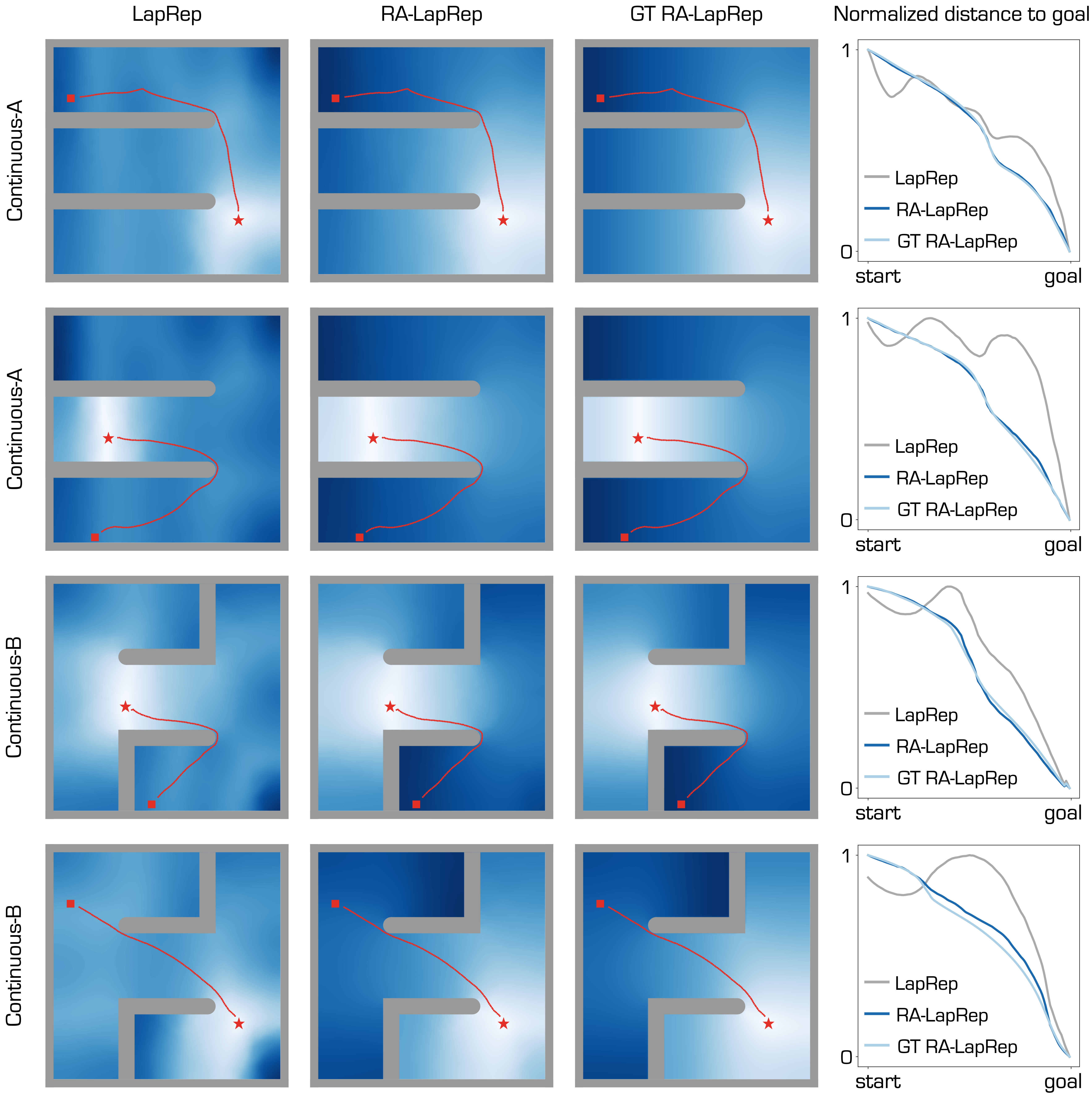}
    \caption{\textbf{Left 3 columns}: Visualization of the Euclidean distance between all states and the goals in continuous environments. For each environment, two additional trajectories (different from the one in the main paper) are shown in red. \textbf{Right}: Normalized distance values for states in the trajectories.}
     \label{fig: supp-dist_continuous}
\end{figure}

\clearpage
\subsection{Reward shaping results with error bars plotted}
\begin{figure}[h]
    \centering
    \includegraphics[width=0.88\linewidth]{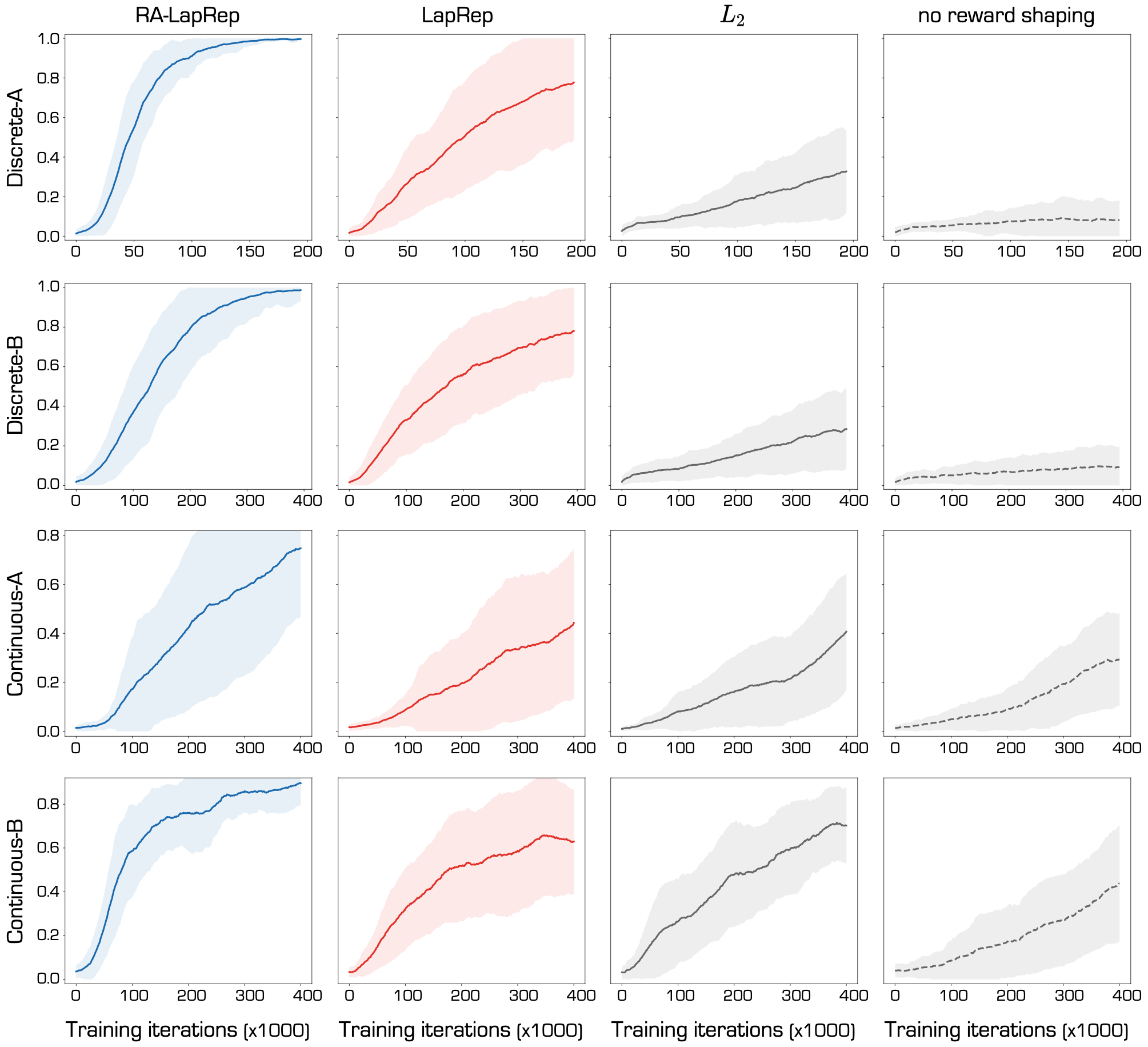}
    \caption{Reward shaping results of different methods, with standard deviation plotted as shaded area.}
     \label{fig: supp-reward_shaping}
\end{figure}

\begin{figure}[h]
    \centering
    \includegraphics[width=0.88\linewidth]{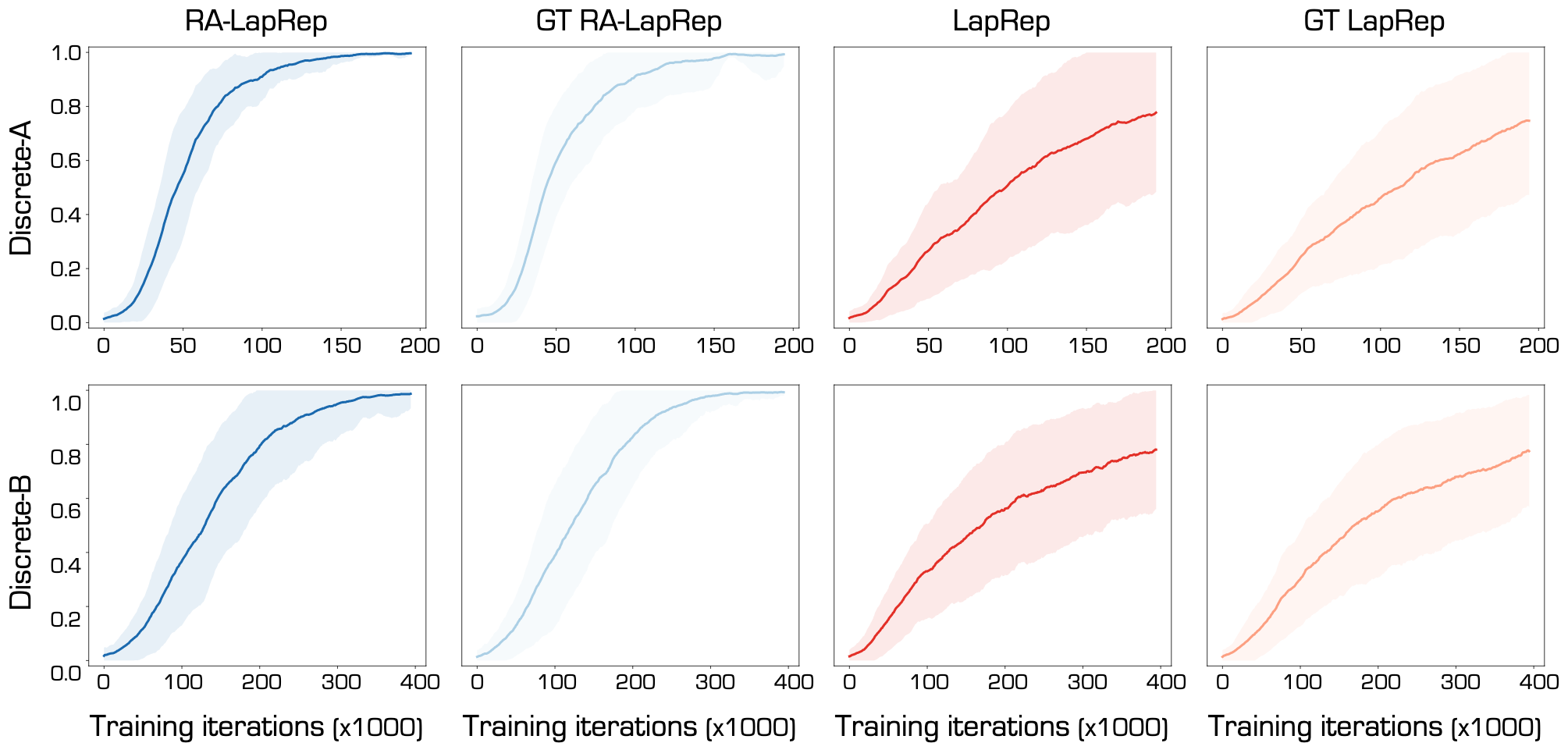}
    \caption{Reward shaping results of using learned or ground truth representations, with standard deviation plotted as shaded area.}
     \label{fig: supp-ablation_gt}
\end{figure}

\clearpage

\begin{figure}[h]
    \centering
    \includegraphics[width=\linewidth]{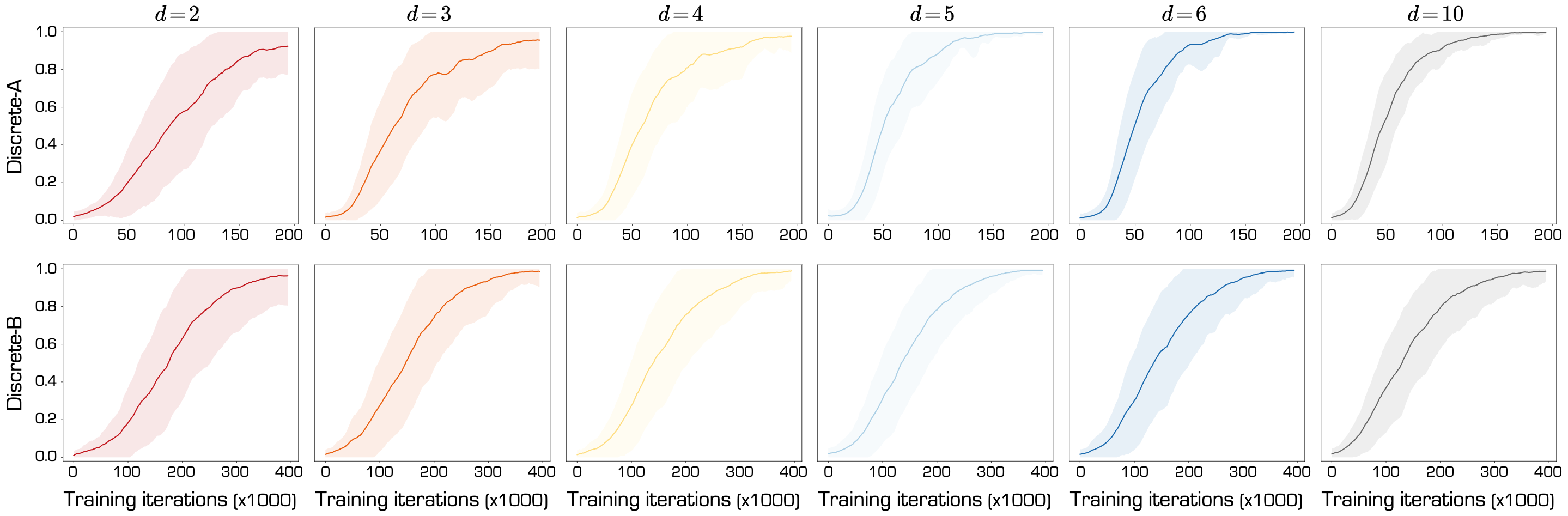}
    \caption{Reward shaping results of using different number of dimensions, with standard deviation plotted as shaded area.}
     \label{fig: supp-ablation_dim}
\end{figure}

\clearpage
\section{Ablation study on uniform state coverage}
\label{sec:ablate-coverage}

\begin{figure}[h]
    \centering
    \includegraphics[width=\linewidth]{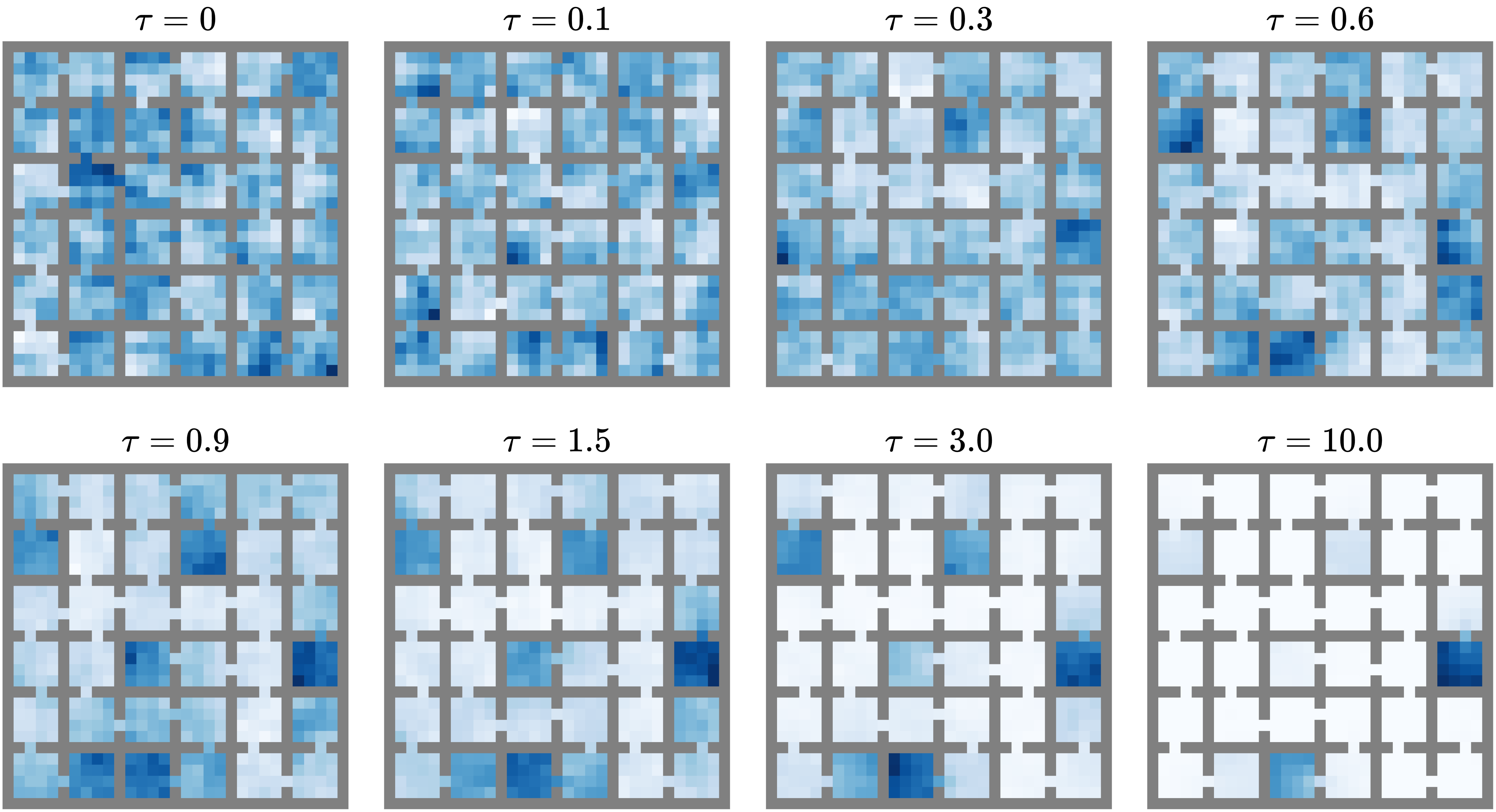}
    \caption{Visualization of the visitation counts of the collected data under different $\tau$.}
     \label{fig: rbtl-visitation}
\end{figure}

We conduct ablative experiments to study the learned \ralaprep would be affected when the
uniformly full state coverage assumption breaks, by manipulating the distribution of the collected data.
For easy comparison, we use \texttt{Discrete-A} environment since it has a visually clear structure.
Specifically, we sample the agent's starting position in each episode from a distribution that can be controlled with a temperature parameter $\tau \in [0, +\infty)$.
When $\tau=0$, it reduces to a uniform distribution. 
As $\tau$ increases, more probability mass will be put on the dead end rooms (see Fig.~\ref{fig: rbtl-visitation}, which visualizes the distribution of the collected data under different $\tau$).

\begin{figure}[h]
    \centering
    \includegraphics[width=\linewidth]{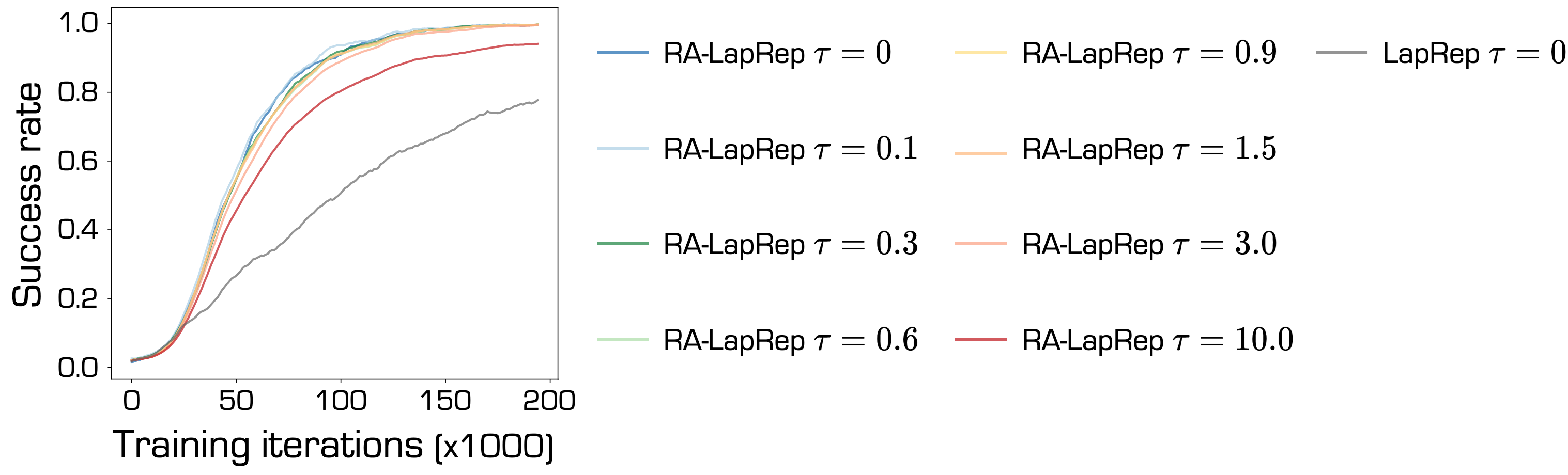}
    \caption{Comparison of the reward shaping results using the learned RA-LapRep under different $\tau$.}
     \label{fig: rbtl-reward_shaping}
\end{figure}

\begin{figure}[h]
    \centering
    \includegraphics[width=\linewidth]{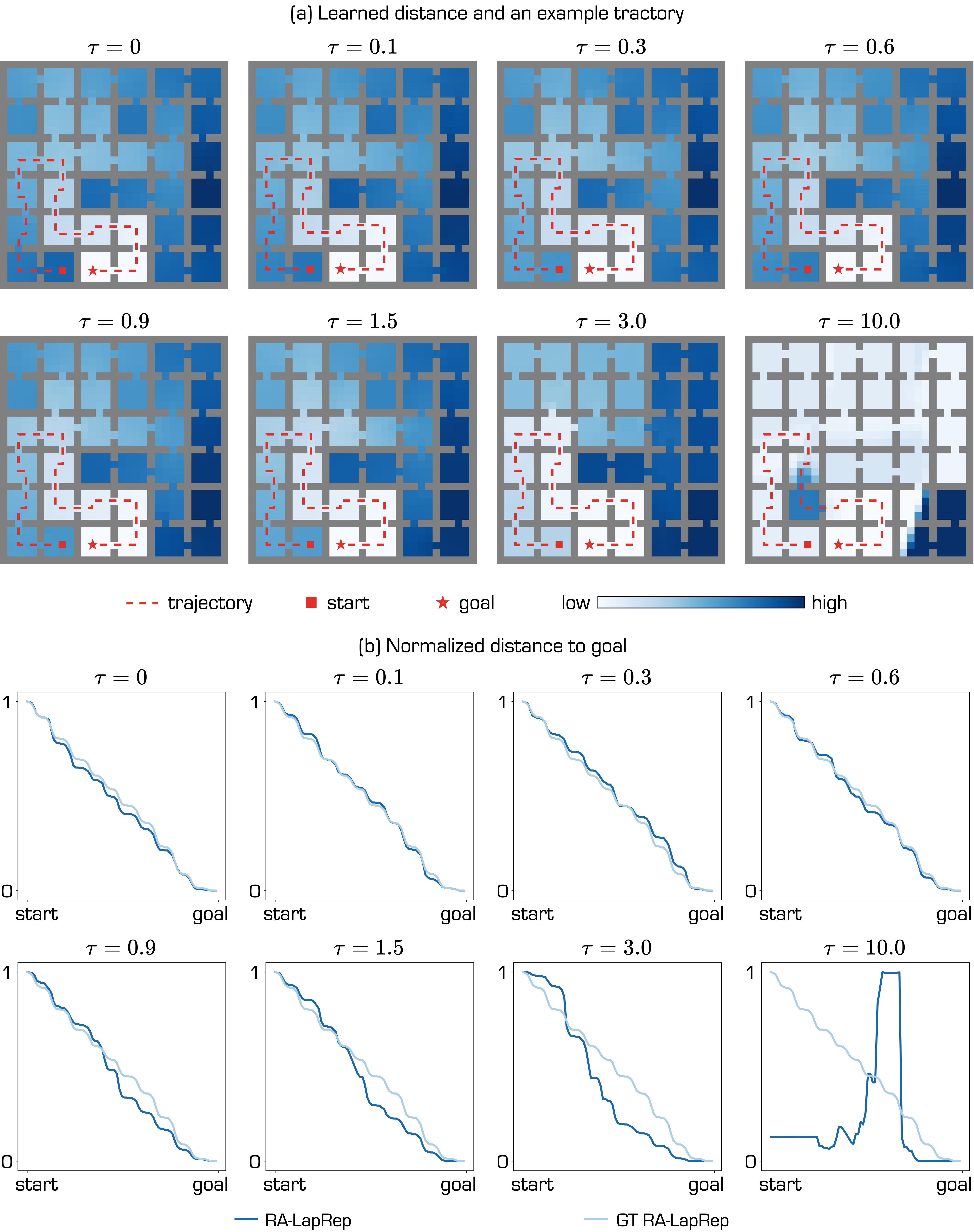}
    \caption{\textbf{(a)}: Visualizations of the Euclidean distances between all states and the goals under learned learned \ralaprep for different $\tau$. Example trajectories are shown in red.
    \textbf{b}: Line charts of the distance values for states in the trajectories (normalized to $[0, 1]$), where the states are sorted by temporal order.}
     \label{fig: rbtl-dist_path_temp}
\end{figure}

For each $\tau \in \{0, 0.1, 0.3, 0.6, 0.9, 1.5, 3.0, 10.0\}$, we learn RA-LapRep with the corresponding collected data.
Note that $\tau=0$ is the setting we used in the main text to learn RA-LapRep.
Hyper-parameters for training are the same as those used in Sec.~\ref{sec:exp-distance}. Similar to Fig.~\ref{fig:dist}, we visualize in Fig.~\ref{fig: rbtl-dist_path_temp} the Euclidean distances under RA-LapRep learned using different $\tau$.
We can see that, for $0 \leqslant \tau \leqslant 3.0 $, the distance is robust to the changes in the data distribution, demonstrating that the RA-LapRep can still capture the inter-state reachability. 
When $\tau$ becomes too large (\eg, $\tau=10$), the method will fail. 
This is because the unvisited area is too large, which results in a disconnected graph and hence violates the assumption behind the methods in \cite{wu2018laplacian, wang2021towards}.
Furthermore, we conduct reward shaping experiments using the learned RA-LapRep under different $\tau$. 
As Fig.~\ref{fig: rbtl-reward_shaping} shows, the performance only drops a little when $\tau$ increases from $0$ to $3.0$. This again shows that our approach is robust to moderate changes in the distribution of the collected data.

\clearpage
\section{Evaluation with top-view image observation}
\label{sec:high-dim}

\begin{figure}[h]
    \centering
    \includegraphics[width=\linewidth]{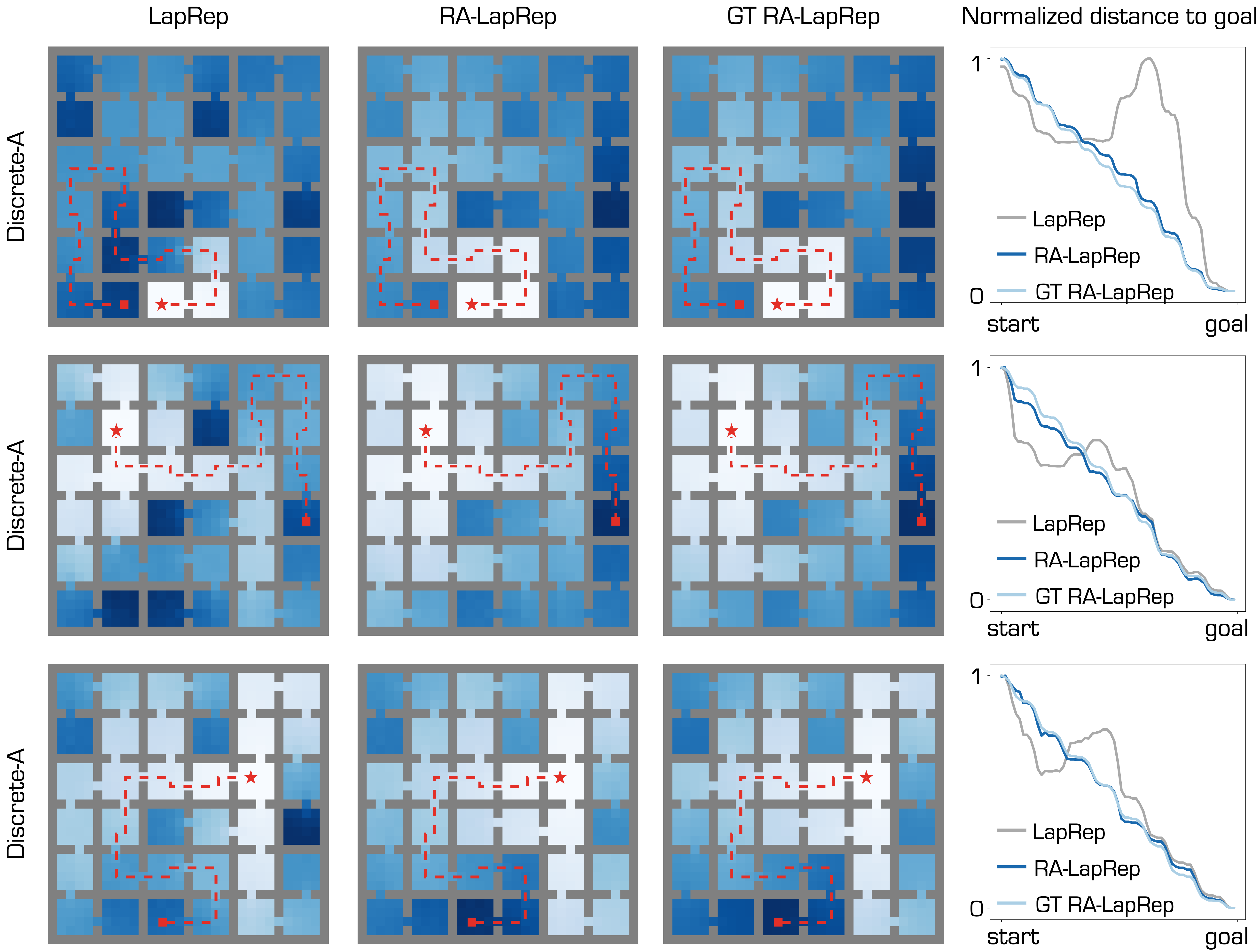}
    \caption{\textbf{Left 3 columns}: Visualization of the Euclidean distance between all states and the goals in \texttt{Discrete-A} environment, when the representations are learned from top-view image observations. \textbf{Right}: Normalized distance values for states in the trajectories.}
     \label{fig: supp-dist_discrete_img-a}
\end{figure}

\begin{figure}[h]
    \centering
    \includegraphics[width=\linewidth]{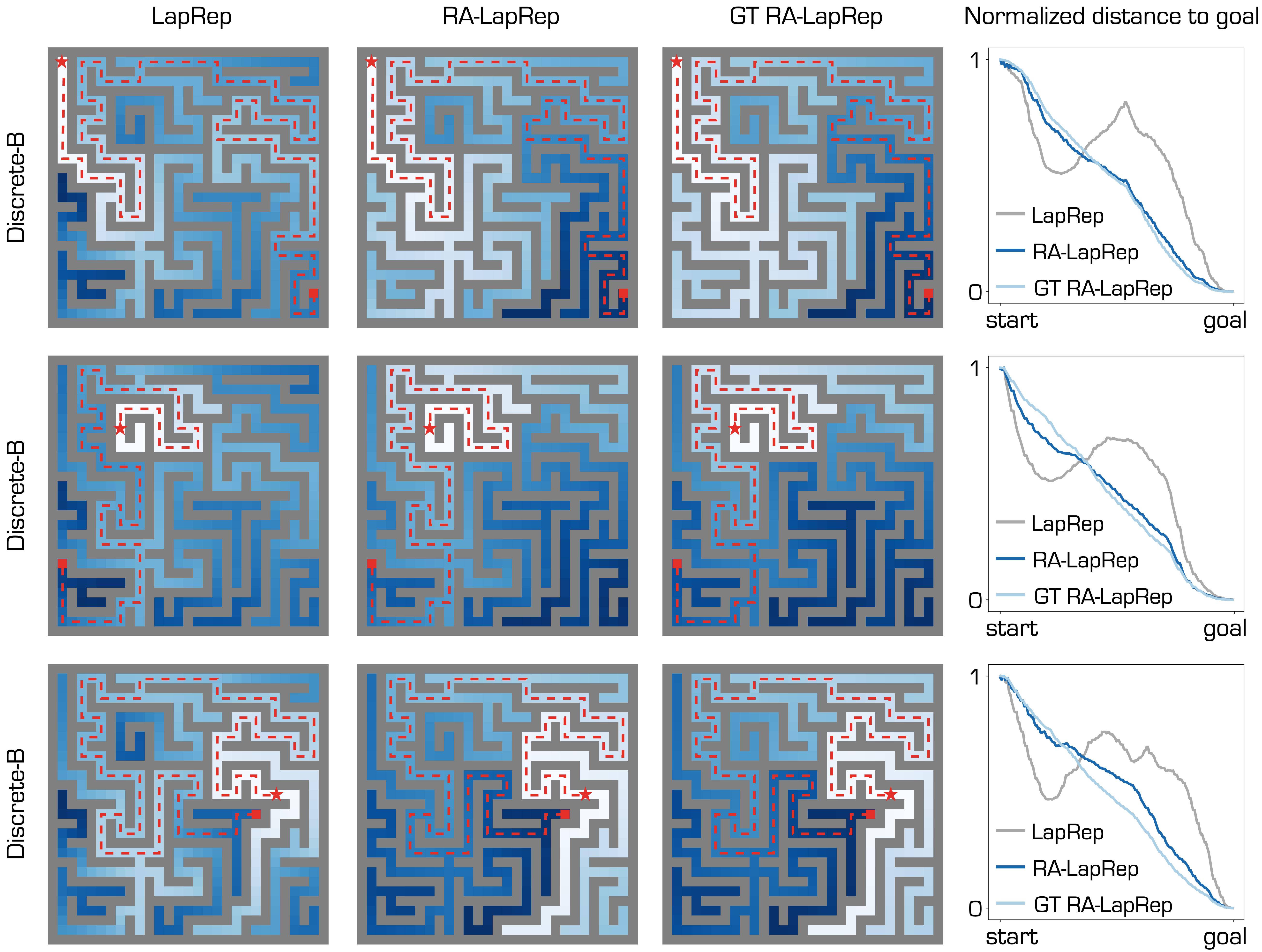}
    \caption{\textbf{Left 3 columns}: Visualization of the Euclidean distance between all states and the goals \texttt{Discrete-B} environment, when the representations are learned from top-view image observations. For each environment, two additional trajectories (different from the one in the main paper) are shown in red. \textbf{Right}: Normalized distance values for states in the trajectories.}
     \label{fig: supp-dist_discrete_img-b}
\end{figure}

\begin{figure}[t]
    \centering
    \includegraphics[width=\linewidth]{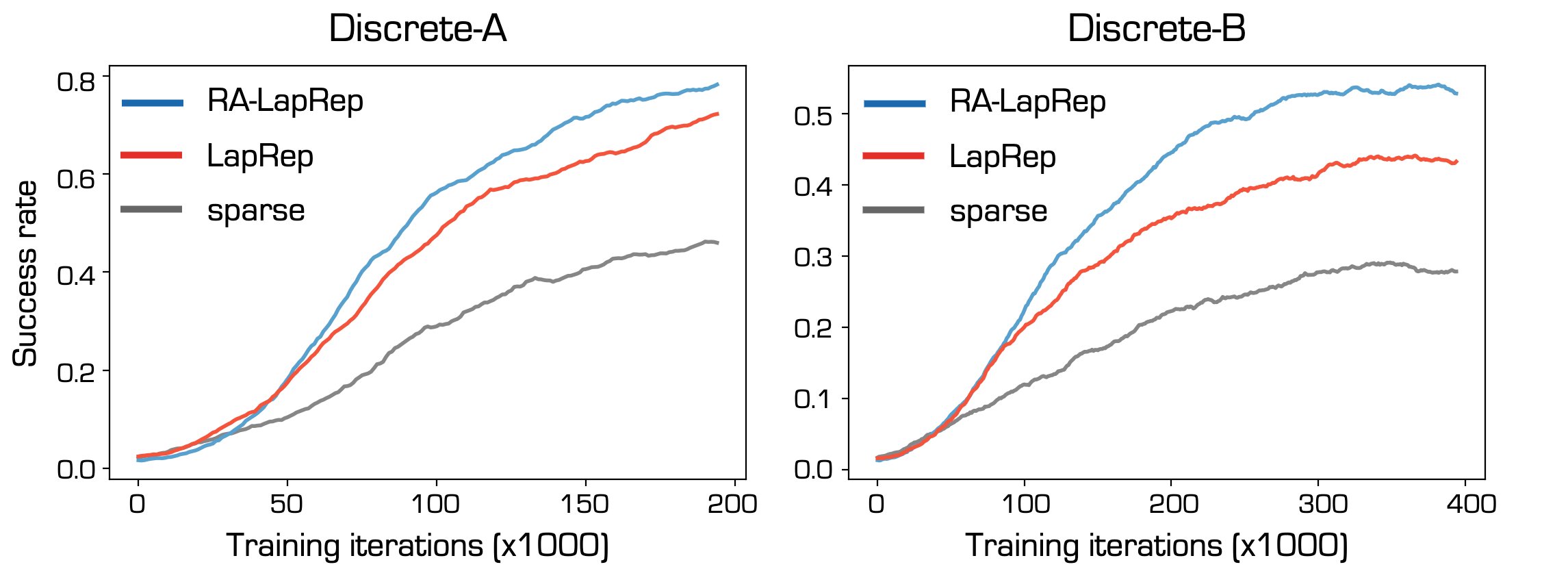}
    \caption{Reward shaping results in goal-reaching tasks using high-dimensional image input.}
     \label{fig:img_reward_shaping}
\end{figure}

We train \laprep and \ralaprep with top-view image observations on two discrete environments. The distances under learned representations are visualized in Figure~\ref{fig: supp-dist_discrete_img-a} and Figure~\ref{fig: supp-dist_discrete_img-b}. The reward shaping results are shown in Figure~\ref{fig:img_reward_shaping}.

\clearpage
\section{Further discussions on related works}
\label{sec:more-related}

Both our work and the reachability network~\citep{savinov2018episodic} view the adjacent states as positive pairs and the distant states as negative pairs, for shaping the representation. The reachability network is trained to classify whether two states are adjacent (within a preset radius) or far away. While being flexible, this method is largely based on intuition. In comparison, our approach is more theoretically grounded and ensures that the distance between two states reflects the reachability between them. As the Figure 11 in \citep{zhang2020generating} shows, even for a very simple two-room environment, the adjacency learned by the reachability network~\citep{savinov2018episodic} is not good (note the adjacency score between $s_1$ and $s_2$).

While the adjacency network~\citep{zhang2020generating} shows some improvements over the reachability network~\citep{savinov2018episodic}, the results are still not satisfying. For example in their Figure 11, the adjacency score between $s_1$ and $s_2$ is lower than the one between $s_1$ and the door connecting two rooms (it is supposed to be higher, since $s_1$ is farther from $s_2$ than from the door). In comparison, our results are verified in more complicated mazes and do not have such issues. More importantly, the adjacency network~\citep{zhang2020generating} needs to maintain an adjacency matrix, which limits its applicability to environments with large or even continuous state spaces.

The dynamic distance method~\citep{hartikainen2020dynamical} also focuses on approximating the commute time between two states. They directly learn a parameterized function to predict the temporal difference between two states sampled from the same trajectory. However, the learned distance is only visualized for a simple S-shaped maze. It is hard to tell whether this method can still learn good representations in environments with more complex geometry. In comparison, our method is able to accurately reflect the inter-state reachability in complex environments (such as Discrete-A and Discrete-B in our paper).

\putbib
\end{bibunit}

\end{document}